\documentclass{article}
\usepackage{iclr2026_conference,times}

\usepackage{amsmath,amsfonts,bm}

\def\eqref#1{equation~\ref{#1}}

\def\1{\bm{1}}

\DeclareMathAlphabet{\mathsfit}{\encodingdefault}{\sfdefault}{m}{sl}
\SetMathAlphabet{\mathsfit}{bold}{\encodingdefault}{\sfdefault}{bx}{n}

\usepackage{marvosym} 
\usepackage{amsmath}
\usepackage{wrapfig}
\usepackage{amssymb}

\usepackage{hyperref}
\usepackage{url}
\usepackage{multirow}
\usepackage{booktabs}
\usepackage{graphicx}
\usepackage{colortbl}
\usepackage{xcolor}
\usepackage{tikz}
\usepackage{adjustbox}
\usepackage{algorithm}
\usepackage{algorithmic}
\usetikzlibrary{positioning,arrows.meta,shapes.geometric,calc}

\title{STVG-R1: Incentivizing Instance-Level \\ Reasoning and Grounding in Videos via \\ Reinforcement Learning}

\iclrfinalcopy 

\author{Xiaowen Zhang$^{1,2}$, Zhi Gao$^{2,3}$, Licheng Jiao$^{1\text{\Letter}}$, Lingling Li$^{1}$, Qing Li$^{2\text{\Letter}}$,\\ 
  \small $^1$Xidian University \small $^2$State Key Laboratory of General Artificial Intelligence, BIGAI, \\
\small $^3$Beijing Institute of Technology  \\ 
\small \url{https://stvg-r1.github.io/}
}

\begin{document}

\maketitle

\begin{abstract}
In vision–language models (VLMs), misalignment between textual descriptions and visual coordinates often induces hallucinations. This issue becomes particularly severe in dense prediction tasks such as spatial–temporal video grounding (STVG). Prior approaches typically focus on enhancing visual–textual alignment or attaching auxiliary decoders. However, these strategies inevitably introduce additional trainable modules, leading to significant annotation costs and computational overhead. In this work, we propose a novel visual prompting paradigm that avoids the difficult problem of aligning coordinates across modalities. Specifically, we reformulate per-frame coordinate prediction as a compact instance-level identification problem by assigning each object a unique, temporally consistent ID. These IDs are embedded into the video as visual prompts, providing explicit and interpretable inputs to the VLMs. Furthermore, we introduce STVG-R1, the first reinforcement learning framework for STVG, which employs a task-driven reward to jointly optimize temporal accuracy, spatial consistency, and structural format regularization. Extensive experiments on six benchmarks demonstrate the effectiveness of our approach. STVG-R1 surpasses the baseline Qwen2.5-VL-7B by a remarkable margin of 20.9\% on m\_IoU on the HCSTVG-v2 benchmark, establishing a new state of the art (SOTA). Surprisingly, STVG-R1 also exhibits strong zero-shot generalization to multi-object referring video object segmentation task, achieving a SOTA 47.3\% $\mathcal{J}\&\mathcal{F}$ on MeViS. 
\end{abstract}

\vspace{-6pt}
\section{Introduction}
\vspace{-6pt}
In video grounding task, hallucination in vision–language models (VLMs) is a common phenomenon, where timestamps may extend beyond video duration or coordinates may exceed the frame resolution \citep{wang2024qwen2, liu2024improved, chen2024longvila}. A widely accepted perspective is that the hallucinations stem from misalignments between the visual and textual modalities \citep{lin2024vila, wang2024cogvlm}. Such misalignment leads to greater performance degradation in dense prediction tasks, where bounding box or segmentation mask for each frame are required. 

\begin{figure}[!t]
\begin{center}
\includegraphics[width=1\textwidth]{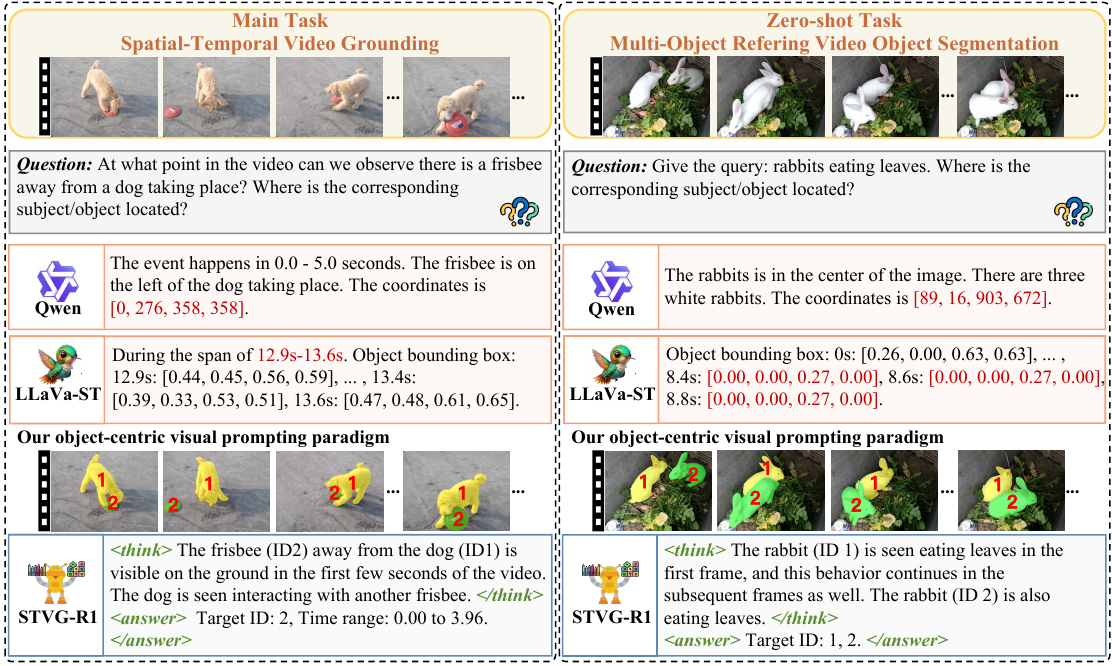}
\end{center}
\setlength{\abovecaptionskip}{2pt}
\caption{Comparisons of general VLMs, specialized VLMs, and proposed STVG-R1 model. While Qwen2.5-VL-7B outputs a single meaningless bounding box without timestamps, LLaVA-ST is restricted to one bounding box per frame. In contrast, STVG-R1 achieves strong performance on both spatial–temporal video grounding and zero-shot multi-object referring video object segmentation.}
\label{fig:fig1}
\end{figure}

To reduce the impact of cross-modal misalignment, existing research focuses on improving the alignment capability of VLMs \citep{wang2025spacevllm, ye2024x} or avoiding direct coordinate prediction \citep{yuan2025sa2va, sun2025sama}. Despite their success, these strategies typically introduce additional learnable components and exhibit limited generalization. As illustrated in Figure~\ref{fig:fig1}, alignment-based approaches \citep{li2025llava} directly output explicit frame-level coordinates, but struggle in multi-object scenes and often yield inconsistent or even meaningless predictions, such as $[0.00,0.00,0.27,0.00]$. In contrast, decoder–based methods alleviate this by introducing segmentation tokens for cross-frame consistent prediction, yet their implicit outputs limit generalization. Motivated by these challenges and prior attempts, we propose a core idea: if the complex per-frame coordinate prediction can be reformulated into a compact and interpretable formulation, it becomes possible to mitigate visual–textual misalignment and enhance generalization.

\setlength{\textfloatsep}{8pt plus 1pt minus 2pt}
\vspace{-3pt}

Based on the above observation, we draw inspiration from existing research on visual prompts, which demonstrates the effectiveness of simple yet consistent referential cues for object representation \citep{shtedritski2023does, cai2024vip, yang2024set}. Taking GPT4Scene \citep{qi2025gpt4scene} as an example, consistent object IDs across multi-view images is embedded to enhance 3D understanding. Following this insight, in this paper, we introduce an object-centric visual prompting paradigm for spatial–temporal video grounding (STVG). Specifically, each object is automatically assigned a unique and temporally consistent identifier throughout the video sequence. Concretely, the first frame is processed with an object detector \citep{tian2025yolov12, liu2024grounding, xiao2023florence} to obtain candidate bounding boxes, which are further refined using the segmentation and tracking capabilities of SAM2 \citep{ravi2024sam2}. To handle newly appearing or previously missed objects, re-detection is performed at fixed intervals, and ReID is employed to maintain temporal consistency. Finally, each candidate instance is overlaid with a numeric marker that serves as its object ID on its center, yielding a compact yet interpretable formulation for video spatial grounding.

Building on this paradigm, we introduce STVG-R1, the first reinforcement learning framework for STVG. Unlike conventional supervised fine-tuning (SFT), which relies on token-level loss, STVG-R1 incorporates a task-driven reward that jointly optimizes temporal accuracy, spatial consistency, and structural correctness. A positive spatial consistency reward is obtained when the predicted object ID is aligned with the ground truth and also falls within the localized temporal segment.

The object-centric visual prompting paradigm achieves substantial performance improvements across four general VLMs in zero-shot settings. Specifically, InternVL3-8B \citep{zhu2025internvl3}, Qwen2.5-VL-7B, Qwen2.5-VL-72B \citep{bai2025qwen2}, and Qwen3-VL-8B improve vIoU@0.3 by +3.6\%, +12.5\%, +6.0\%, and +28.3\% on HCSTVG-v1 \citep{tang2021human}. Beyond zero-shot scenarios, the enhanced reasoning capability introduced by reinforcement learning establishes new SOTA on five benchmarks. Remarkably, the STVG-R1 also achieves strong performance on the unseen multi-object referring video object segmentation task, highlighting its robust generalization ability. We attribute this generalization to the object-centric visual prompts, which provide explicit object identifiers during reinforcement learning, enabling instance-level reasoning and grounding.

The main contributions of this paper are as follows: (1) We introduce a simple yet effective object-centric visual prompting paradigm that reformulates dense per-frame coordinate prediction into a compact object ID identification task. (2) We propose STVG-R1, the first reinforcement learning framework for spatial–temporal video grounding, built upon the GRPO algorithm. (3) Extensive experiments across six benchmarks demonstrate the effectiveness of our approach. Moreover, its strong performance on the unseen multi-object referring video object segmentation task further highlights its generalization capability.

\section{Related Work}
\begin{figure}[!t]
\begin{center}
\includegraphics[width=1\textwidth]{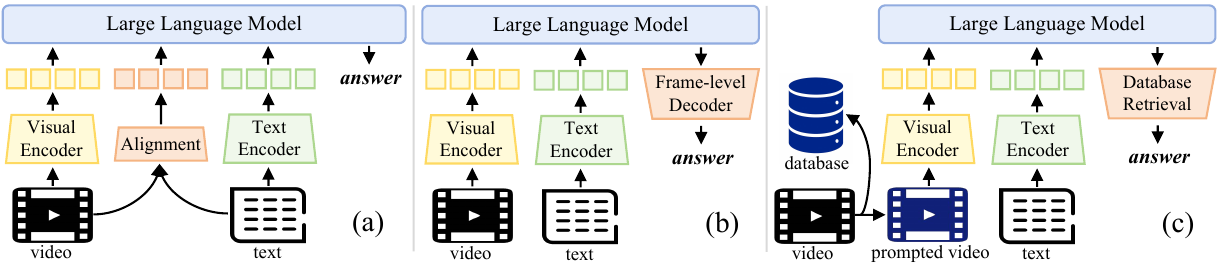}
\end{center}
\caption{Comparison of paradigms: (a) VLM produces both timestamps and frame-level coordinates with a \textbf{trainable} alignment block; (b) VLM generates segmentation tokens, which are then processed by a \textbf{trainable} decoder; (c) our method uses \textbf{training-free} object-centric visual prompted video for spatial-temporal video grounding.}
\label{fig:fig2}
\end{figure}

\vspace{-6pt}
\subsection{Spatial Temporal Video Grounding}
\vspace{-6pt}
In the research on spatial–temporal video grounding, existing approaches can be broadly categorized into models based on visual–language pretraining (VLP) and models leveraging VLMs. VLP-based methods typically employ pretrained encoders, such as CLIP \citep{radford2021learning}, I3D \citep{carreira2017quo}, InternVideo-v2 \citep{wang2024internvideo2}, and BERT \citep{devlin2019bert}, to extract visual and textual features, followed by the design of task-specific modules for multimodal feature fusion and tailored decoding. These approaches \citep{gu2024context, gu2025knowing, lin2022stvgformer} still demonstrate dominant performance on several STVG benchmarks \citep{tang2021human, zhang2020does}. However, despite their effectiveness, these VLP-based task-specific models continue to struggle with generalization, even on simpler spatial-only or temporal-only video grounding tasks.

Recent efforts have increasingly adopted VLMs \citep{li2024llava, bai2025qwen2, abdin2024phi, zhang2024video} for video spatial grounding, owing to their superior cross-modal reasoning and generalization abilities. Within this line of research, as shown in Figure~\ref{fig:fig2}(a), one direction directly exploits VLMs for dense prediction, producing both temporal segments and frame-level spatial localization results. For example, LLaVA-ST \citep{li2025llava} enhances the alignment between textual descriptions and visual coordinates by incorporating additional tokens into the input text embeddings. Subsequently, SpaceVLLM \citep{wang2025spacevllm} follows a similar strategy by introducing spatial-temporal query tokens to address the alignment challenge. However, these additional trainable tokens require large-scale, high-quality dense prediction data and lead to substantial computational overhead. As shown in Figure~\ref{fig:fig2}(b), another direction mitigates the impact of misalignment by prompting VLMs to generate segmentation tokens \citep{yuan2025sa2va, sun2025sama,munasinghe2025videoglamm}, which are then passed into a trainable decoder \citep{ravi2024sam2}. However, their reliance on iterative decoding further increases training complexity and time.

\vspace{-6pt}
\subsection{Reinforcement Learning in VLMs}
\vspace{-6pt}
Reinforcement learning (RL) has demonstrated strong potential in improving the reasoning capabilities of LLMs, particularly through reinforcement learning with verifiable reward (RLVR) \citep{guo2025deepseek, chen2025r1v, jaech2024openai}. For VLMs, many works \citep{liu2025visual, shen2025vlm, zhang2025r1, chen2025scaling,zhang2025chain} also apply this reward-driven training paradigm to tackle complex tasks \citep{yang2025thinking, fu2025video}. Recent studies further extend this direction to video understanding and multimodal agents~\citep{fan2024videoagent, fan2025embodied}. More closely related to our setting, Video-R1 \citep{feng2025video} is the first attempt to explore the R1 paradigm in the video domain, introducing the T-GRPO algorithm to explicitly encourage temporal understanding by shuffling the order of input video frames. Building on this foundation, Time-R1 \citep{wang2025time} proposes a novel post-training framework for temporal video grounding, also based on the Group Relative Policy Optimization (GRPO) algorithm \citep{shao2024deepseekmath}. More encouragingly, Time-R1 demonstrates that using continuous metrics such as IoU as rewards provides more intuitive optimization signals and achieves better performance than token-level supervised fine-tuning. However, applying RL to jointly address spatial–temporal video grounding remains an underexplored yet promising direction.To bridge this gap, STVG-R1 integrates GRPO with STVG-specific rewards, achieving superior performance.

\begin{figure}[t]
\begin{center}
\includegraphics[width=1\textwidth]{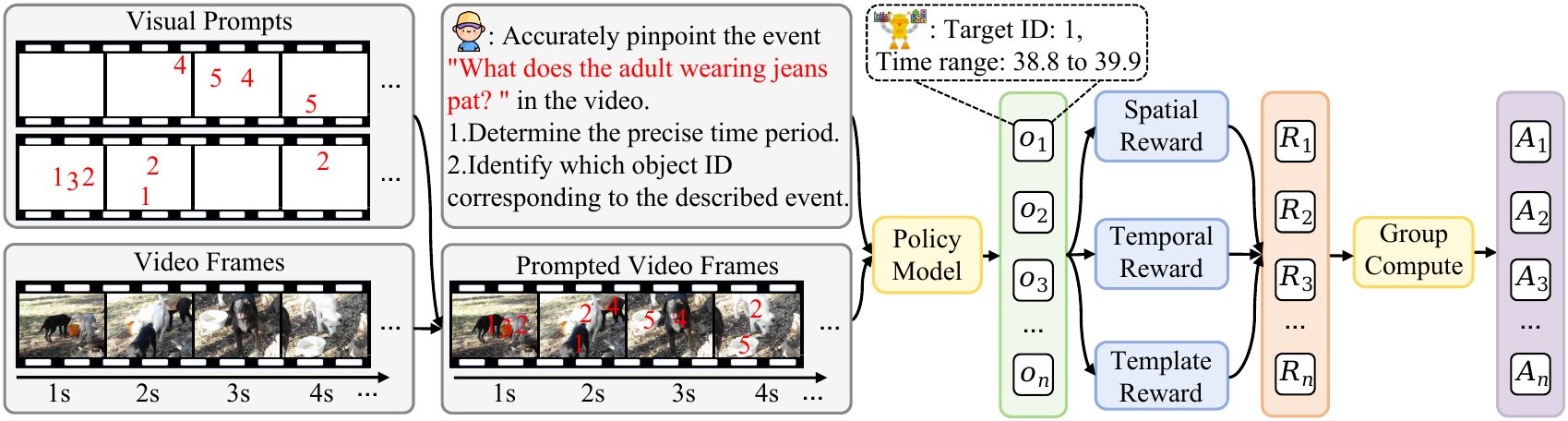}
\end{center}
\caption{An illustration of our proposed STVG-R1 framework. Each object is assigned a unique ID via visual prompts, and the policy model is trained with spatial, temporal, and template rewards.}
\label{fig:stvg_r1_pipeline}
\end{figure}

\section{Method}
\label{sec:method}
\subsection{STVG-R1 Framework}
Our approach reformulates spatial–temporal video grounding as a paradigm shift from dense per-frame bounding box regression to a compact formulation based on visual prompts. Figure~\ref{fig:stvg_r1_pipeline} illustrates the overall architecture of the STVG-R1 model. Specifically, given a video $\mathcal{V}=\{I_1,\ldots,I_T\}$ with $T$ frames, each frame $I_t$ is first augmented with a set of visual prompts,
\begin{equation}
\mathcal{P}_t=\{p_1^{t},\ldots,p_{K_t}^{t}\},\qquad 
\tilde I_t \triangleq I_t \oplus \mathcal{P}_t,
\end{equation}
where $t\in\{1,\ldots,T\}$ indexes frames, $K_t$ is the number of instances in frame $t$, and $\oplus$ denotes overlaying the visual prompts onto the frame $I_t$, yielding the augmented sequence
$\tilde{\mathcal{V}}=\{\tilde I_1,\ldots,\tilde I_T\}$. 

To control memory consumption, we constrain each video to approximately $R=1.6\times 10^6$ pixels in total. Concretely, for a video $\mathcal{V}$ with frame resolution $H\times W$ and duration $D$ seconds, we resize frames to $H'\times W'\approx R/(2D)$, where frames are uniformly sampled at 2 FPS. For example, a 30-second video yields 60 frames, each with a resolution of about $96\times96\times3$. Finally, the sequence of visual prompt–augmented frames $\tilde{\mathcal{V}}$ and a textual query $q$ are fed into a VLM $\pi_\theta$, which jointly predicts the temporal interval $[t_s,t_e]$ and the corresponding object identifier $\imath$.

\subsection{Object-Centric Prompted Video Construction}
\textbf{Data format.} Each sample is denoted as $\{\mathcal{V}, q, P, M, A\}$, where $P=\{\mathcal{P}_t\}_{t=1}^T$ represents the set of visual prompts over frames, $M$ is the segmentation mask database, and $A$ is the ground-truth answer, defined as the target object ID. Concretely, $M$ stores for each frame $I_t$ a set of instance IDs paired with their run-length encoded masks \citep{golomb1966run}. For consistency with the ground-truth annotations, each mask $m_k^t$ is further converted into its corresponding bounding box $b_k^t$.

To formally derive $A$, we establish a frame-level correspondence within the ground truth temporal interval. For each frame $I_t$, we compute the IoU between the ground-truth $g_t$ and all candidate bounding boxes $\{b_{k}^t\}_{k=1}^{K_t}$, and assign to frame $t$ the ID ${\imath}_t$ with the highest overlap:
\begin{equation}
{\imath}_t = \arg\max_{k \in \{1,\ldots,K_t\}} \mathrm{IoU}(g_t, b_{k}^t).
\end{equation}
Over the entire video $\mathcal{V}$, the final answer $A$ is obtained by majority voting:
\begin{equation}
A = \arg\max_{i} \ \sum_{t=1}^T \mathbf{1}[{\imath}_t = i],
\end{equation}
where $\mathbf{1}[\cdot]$ denotes the indicator function.  
This defines the target object as the identifier with the highest overall IoU consistency across the video.

\textbf{Data generation pipeline.}
To construct object-centric visual prompted videos, we integrate several existing vision models into a unified pipeline. The first frame $I_1$ of each video is processed by an off-the-shelf object detector to produce bounding boxes $\{b_{k}^1\}_{k=1}^{K_1}$ for all candidate instances across COCO categories. These detections serve as prompts for SAM2, which generates high-quality segmentation masks $\{m_{k}^1\}_{k=1}^{K_1}$ that are then propagated to subsequent frames via tracking. To capture newly appearing or previously missed objects, periodic re-detection with IoU-based matching is performed, where each detection is compared against the masks already tracked to the same frame and is treated as a new instance only when its geometric overlap with all existing objects remains consistently low. Once a new instance is identified, SAM2 is further applied to perform both forward and backward tracking from the discovery frame to recover its complete temporal trajectory. For each mask $m_k^t$, we embed a compact visual prompt $p_k^t$ at its centroid $(x_k^t, y_k^t)$. 

Importantly, although the COCO taxonomy does not fully cover all categories present in the videos, the detector can still provide bounding boxes for nearly all instances. For example, while fish is absent from the COCO label set, such instances are nevertheless detected under alternative categories. These semantic misclassifications do not affect our framework, as supervision depends only on consistent instance identities rather than precise class labels.

\textbf{Data source.} Two widely adopted STVG datasets are used for training. HCSTVG \citep{tang2021human} focuses on human-centric grounding data. We merge the training splits of v1 and v2, and remove any samples that appear in the validation or test sets. VidSTG \citep{zhang2020does} covers both humans and objects with diverse query types, providing both visual and linguistic diversity.

\textbf{Preprocessing Pipeline Robustness Analysis.}
The reliability of the object-centric visual prompting pipeline is critical for stable reinforcement learning. We analyze potential failure modes in the object-centric prompting pipeline that could disrupt consistent instance identity construction. Global detection failures, defined as cases where the target object is not detected in any frame of the video, occur in fewer than 1\% of all samples, indicating that the majority of target objects can be detected and assigned instance IDs. To address local missing detections caused by occlusion or fast motion, the pipeline incorporates periodic re-detection and SAM2’s bidirectional propagation to recover full object trajectories. Identity consistency is further reinforced via majority voting during ID assignment, while a lightweight ID-repair step at evaluation time resolves occasional re-identification inconsistencies, as detailed in Appendix~\ref{app:IDrepair}. Together, these mechanisms mitigate detection and tracking noise, yielding stable instance identities for downstream optimization.

\subsection{Enhancing VLMs with Reinforcement Learning}
Since dense per-frame prediction is reformulated as a compact instance-level identification task, reinforcement learning can be directly applied to optimize the policy model with task-specific rewards. These identifiers further enable the model to produce more precise and interpretable reasoning chains during RL training, leading to more coherent spatial–temporal predictions.

\textbf{Reward modeling.} Building on DeepSeek-R1~\citep{guo2025deepseek}, the reward design in STVG-R1 integrates both accuracy and format components. The accuracy reward measures the correctness of predictions, while the format reward enforces structural compliance with a predefined reasoning template. To capture both temporal and spatial accuracy, the accuracy reward is further decomposed into a temporal IoU reward and a spatial consistency reward. The temporal IoU reward quantifies the overlap between the predicted interval $[t_s,t_e]$ and the ground-truth segment $[t'_s,t'_e]$, defined as:
\begin{align}
r_{t}(o) = \frac{[t_s, t_e] \cap [t'_s, t'_e]}{[t_s, t_e] \cup [t'_s, t'_e]},
\end{align}
where $A \cap B$ and $A \cup B$ denote the intersection and union of intervals $A$ and $B$, respectively.

The spatial consistency reward verifies whether the predicted object ID is correct and appears within the localized temporal segment:
\begin{align}
r_{s}(o) = 
\begin{cases}
1, & \text{if $\imath = \imath^{*}$ and $\imath$ appears in } [t_s,t_e], \\
0, & \text{otherwise,}
\end{cases}
\end{align}
where $\imath$ and $\imath^{*}$ denote the predicted and ground-truth object ID. This design is consistent with the vIoU metric in STVG, defined as $\lvert P_u\rvert^{-1}\!\sum_{t\in P_i}\operatorname{IoU}(b_t, b_t^{*})$, where $P_i$ and $P_u$ are the intersection and union of the predicted and ground-truth temporal segments, and $b_t$ and $b_t^{*}$ are the predicted and ground-truth bounding boxes at frame $t$. Since vIoU jointly evaluates temporal and spatial accuracy, constraining the predicted ID to fall within the localized segment prevents trivial solutions and mitigates overfitting to dataset-specific temporal patterns, improving optimization stability.

Beyond accuracy, the format reward $r_{f}(o)$ enforces compliance with the predefined reasoning structure, encouraging the model to explicitly generate its reasoning process. A value of 1 is assigned only if the response encloses the reasoning within \verb|<think>...</think>| and the final prediction within \verb|<answer>...</answer>|. Reasoning traces with timestamps and instance IDs provide clearer references and more precise grounding.

The overall reward is the sum of the three components:
\begin{align}
R(o) = r_{t}(o) + r_{s}(o) + r_{f}(o).
\label{eq:reward}
\end{align}

\subsection{Training Strategies}
The training objective is to optimize the policy model $\pi_\theta$ with GRPO \citep{guo2025deepseek}, which has shown strong effectiveness in tasks with well-defined evaluation signals. Given a video–text query pair $(\tilde{\mathcal{V}}, q)$ sampled from the training distribution $\mathcal{D}$, the model generates $n$ candidate responses $o=\{o_1,\dots,o_n\}$, each assigned a reward $R(o_i)$ as defined in Equation~\ref{eq:reward}. In our experiments, $n$ is set to 8. To normalize rewards within a group, the advantage $A_i$ of each response is computed as:
\begin{align}
A_i = \frac{R(o_i)-\text{mean}(\{R(o_j)\}_{j=1}^n)}{\text{std}(\{R(o_j)\}_{j=1}^n)}.
\end{align}

The policy update objective encourages the current policy $\pi_\theta$ to assign higher probabilities to responses with larger normalized advantages, relative to the previous policy $\pi_{\theta_{\text{old}}}$. Formally, the optimization objective is defined as:
\begin{align}
&\mathcal{J}_{\text{GRPO}}(\theta) 
= \mathbb{E}_{(\tilde{\mathcal{V}}, q) \sim \mathcal{D},\, \{o_i\}_{i=1}^n \sim \pi_{\theta_{\text{old}}}(\cdot|q)} \nonumber \\
&\quad \Bigg[
\frac{1}{n}\sum_{i=1}^n
\Big(
\min
\Big(
\frac{\pi_\theta(o_i|q)}{\pi_{\theta_{\text{old}}}(o_i|q)} A_i,\,
\mathrm{clip}\Big(\frac{\pi_\theta(o_i|q)}{\pi_{\theta_{\text{old}}}(o_i|q)}, 1-\epsilon, 1+\epsilon\Big) A_i
\Big) 
- \beta D_{\mathrm{KL}}(\pi_\theta \| \pi_{\mathrm{ref}})
\Big)
\Bigg],
\end{align}
where $\epsilon$ is the clipping parameter, $\beta$ controls the strength of KL regularization, and $\pi_{\mathrm{ref}}$ denotes the frozen reference policy. The clipping term prevents excessively large updates, while the KL penalty constrains policy drift, together stabilizing optimization.

\section{Experiments}
\subsection{Setting} 
\label{sec:setting}
\textbf{Implementation details.} The object detector used during our training and evaluation is YOLOv12-x \citep{tian2025yolov12} with a confidence threshold of 0.25, and SAM2.1-large is the segmentation and tracking model. Re-detection is performed every 15 frames, where a detection is treated as a new instance only if its IoU with all tracked objects is below 0.4 and its overlap ratio with all tracked objects is below 0.6. We employ the Qwen2.5-VL-7B \citep{bai2025qwen2} as the pre-trained model. We use AdamW \citep{loshchilov2017decoupled} optimizer with a linear learning rate scheduler. The learning rate is $1.0e-6$ and the batch size is 1 per device. The model is trained for 1 epoch on our object-centric visual prompting dataset, and all experiments are conducted on 8$\times$A100 GPUs.

\textbf{Benchmarks.} For STVG, HCSTVG-v1 and HCSTVG-v2 \citep{tang2021human} are widely used for human-centric grounding, while ST-Align \citep{li2025llava} extends evaluation to both humans and objects and supports spatial video grounding (SVG). To capture fine-grained spatial understanding, MeViS \citep{ding2023mevis} evaluates mask-level grounding under complex multi-object scenarios. Beyond STVG and SVG, we also consider video temporal grounding (VTG) with Charades-STA \citep{gao2017tall} and TVGBench \citep{wang2025time} to evaluate generalization.

\textbf{Evaluation metrics.} For STVG, following \citep{yang2022tubedetr, gu2024context}, we report m\_tIoU for temporal localization accuracy and m\_vIoU, vIoU@R for joint spatial–temporal grounding quality. For VTG, we adopt m\_tIoU and tIoU@R. For mask-level referring video object segmentation, $\mathcal{J}$ is used to assess region similarity and $\mathcal{F}$ to measure contour accuracy.

\setlength{\tabcolsep}{3pt}
\begin{table*}[!t]
\centering
\renewcommand{\arraystretch}{0.9}
\caption{Performance comparison with state-of-the-art models on HCSTVG-v1 \textit{test} set and HCSTVG-v2 \textit{val} set (\%). The results of GroundingGPT-7B are reported from SpaceVLLM, while those of InternVL3-8B, Qwen2.5-VL-7B, Qwen2.5-VL-72B and Qwen3-VL-8B are generated by our experiments. The best and second-best results are shown in \textbf{bold} and \underline{underlined}.}
\resizebox{\textwidth}{!}{
\begin{tabular}{l|cccc|cccc}
\toprule
\multirow{2}{*}{Models} & \multicolumn{4}{c|}{HCSTVG-v1} & \multicolumn{4}{c}{HCSTVG-v2} \\
\cmidrule(lr){2-5} \cmidrule(lr){6-9}
& m\_tIoU & m\_vIoU & vIoU@0.3 & vIoU@0.5 & m\_tIoU & m\_vIoU & vIoU@0.3 & vIoU@0.5 \\ \midrule 
TubeDETR & - & 32.4 & 49.8 & 23.5 & 53.9 & 36.4 & 58.8 & 30.6 \\
STVGFormer & - & 36.9 & 62.2 & 34.8 & 58.1 & 38.7 & 65.5 & 33.8 \\
CG-STVG & 52.8 & 38.4 & 61.5 & 36.3 & 60.0 & 39.5 & 64.5 & 36.3 \\
TA-STVG & 53.0 & \underline{39.1} & 63.1 & 36.8 & \underline{60.4} & \underline{40.2} & \underline{65.8} & \underline{36.7} \\
\midrule
GroundingGPT-7B & 22.2 & 16.7 & 15.0 & 4.9 & 19.6 & 14.7 & 16.6 & 3.1 \\
SpaceVLLM-7B & \textbf{56.9} & \textbf{39.3} & \underline{66.6} & \underline{36.9} & 58.0 & 34.0 & 56.9 & 24.7 \\
InternVL3-8B & 22.6 & 11.7 & 15.3 & 2.8 & 24.9 & 12.8 & 14.2 & 3.2 \\
\rowcolor{gray!8}
\hspace*{2em}+\textit{VisualPrompt} & 22.9 \textcolor{red!80!black}{$\uparrow$+0.3} & 13.8 \textcolor{red!80!black}{$\uparrow$+2.1} & 18.9 \textcolor{red!80!black}{$\uparrow$+3.6} & 4.2 \textcolor{red!80!black}{$\uparrow$+1.4} & 25.0 \textcolor{red!80!black}{$\uparrow$+0.1} & 15.4 \textcolor{red!80!black}{$\uparrow$+2.6} & 18.1 \textcolor{red!80!black}{$\uparrow$+3.9} & 4.5 \textcolor{red!80!black}{$\uparrow$+1.3} \\
Qwen2.5-VL-7B & 40.3 & 19.7 & 28.2 & 7.9 & 45.1 & 19.3 & 26.0 & 8.2 \\ 
\rowcolor{gray!8}
\hspace*{2em}+\textit{VisualPrompt} & 38.7 \textcolor{green!80!black}{$\downarrow$-1.6} & 24.8 \textcolor{red!80!black}{$\uparrow$+5.1} & 40.7 \textcolor{red!80!black}{$\uparrow$+12.5} & 13.4 \textcolor{red!80!black}{$\uparrow$+5.5} & 44.7 \textcolor{green!80!black}{$\downarrow$-0.4} & 19.5 \textcolor{red!80!black}{$\uparrow$+0.2} & 28.7 \textcolor{red!80!black}{$\uparrow$+2.7} & 10.9 \textcolor{red!80!black}{$\uparrow$+2.7} \\
Qwen2.5-VL-72B & 40.7 & 23.9 & 37.0 & 15.1 & 43.9 & 23.4 & 36.1 & 13.4 \\
\rowcolor{gray!8}
\hspace*{2em}+\textit{VisualPrompt} & 38.8 \textcolor{green!80!black}{$\downarrow$-1.9} & 26.0 \textcolor{red!80!black}{$\uparrow$+2.1} & 43.0 \textcolor{red!80!black}{$\uparrow$+6.0} & 15.2 \textcolor{red!80!black}{$\uparrow$+0.1} & 42.0 \textcolor{green!80!black}{$\downarrow$-1.9} & 27.3 \textcolor{red!80!black}{$\uparrow$+3.9} & 43.5 \textcolor{red!80!black}{$\uparrow$+7.4} & 16.3 \textcolor{red!80!black}{$\uparrow$+2.9} \\
Qwen3-VL-8B & 48.6 & 19.5 & 28.2 & 10.6 & 51.2 & 17.2 & 22.8 & 7.6 \\
\rowcolor{gray!8}
\hspace*{2em}+\textit{VisualPrompt} & 48.7 \textcolor{red!80!black}{$\uparrow$+0.1} & 33.0 \textcolor{red!80!black}{$\uparrow$+13.5} & 56.5 \textcolor{red!80!black}{$\uparrow$+28.3} & 27.3 \textcolor{red!80!black}{$\uparrow$+16.7} & 52.2 \textcolor{red!80!black}{$\uparrow$+1.0} & 35.8 \textcolor{red!80!black}{$\uparrow$+18.6} & 57.9 \textcolor{red!80!black}{$\uparrow$+35.1} & 30.3 \textcolor{red!80!black}{$\uparrow$+22.7} \\
\midrule 
\rowcolor{gray!15}
STVG-R1 & \textbf{56.9} & \underline{39.1} & \textbf{66.7} & \textbf{38.6} & \textbf{61.3} & \textbf{40.8} & \textbf{67.9} & \textbf{38.8} \\
\bottomrule
\end{tabular}
}
\label{table:hcstvg}
\end{table*}
\setlength{\tabcolsep}{6pt}

\setlength{\tabcolsep}{4pt}
\begin{table*}[!t]
\centering
\renewcommand{\arraystretch}{0.9}
\caption{Performance comparison with state-of-the-art models on ST-Align benchmark (\%). The results of Qwen2.5-VL-7B are generated by our experiments.}
\resizebox{\textwidth}{!}{
\begin{tabular}{l|cccc|ccc}
\toprule
\multirow{2}{*}{Models} & \multicolumn{4}{c|}{Spatial-Temporal Video Grounding} & \multicolumn{3}{c}{Video Spatial Grounding}\\
\cmidrule(lr){2-5} \cmidrule(lr){6-8}
& tIoU@0.5 & m\_tIoU & vIoU@0.5 & m\_vIoU & vIoU@0.3 & vIoU@0.5 & m\_vIoU \\ \midrule
GroundingGPT-7B & 7.1 & 12.2 & 2.9 & 9.2 & 19.7 & 5.4 & 17.9 \\
VTimeLLM-7B & 7.1 & 15.5 & - & - & - & - & - \\
Grounded-VideoLLM-7B & 30.0 & 33.0 & - & - & - & - & - \\
LLava-ST-7B & \textbf{44.6} & \underline{43.8} & \underline{21.1} & \underline{22.8} & \underline{47.2} & \underline{30.9} & \underline{32.5} \\
Qwen2.5-VL-7B & 35.2 & 37.4 & 17.1 & 14.3 & 44.6 & 39.5 & 35.5 \\ 
\rowcolor{gray!8}
\hspace*{2em}+\textit{VisualPrompt} & 36.0 \textcolor{red!80!black}{$\uparrow$+0.8} & 38.3 \textcolor{red!80!black}{$\uparrow$+0.9} & 21.5 \textcolor{red!80!black}{$\uparrow$+4.4} & 19.5 \textcolor{red!80!black}{$\uparrow$+5.2} & 57.7 \textcolor{red!80!black}{$\uparrow$+13.1} & 51.4 \textcolor{red!80!black}{$\uparrow$+11.9} & 46.6 \textcolor{red!80!black}{$\uparrow$+11.1} \\ 
\midrule
\rowcolor{gray!15}
STVG-R1 & \underline{43.6} & \textbf{45.1} & \textbf{25.9} & \textbf{23.4} & \textbf{60.3} & \textbf{53.9} & \textbf{48.6} \\
\bottomrule
\end{tabular}
}
\label{table:STalign}
\end{table*}
\setlength{\tabcolsep}{6pt}

\subsection{Evaluation Results on Spatial Temporal Video Grounding}
\label{sec:stvg}
Table~\ref{table:hcstvg} and Table~\ref{table:STalign} present results on HCSTVG-v1/v2 and ST-Align. TubeDETR \citep{yang2022tubedetr}, STVGFormer \citep{lin2023collaborative}, CG-STVG \citep{gu2024context}, and TA-STVG \citep{gu2025knowing} are four VLP-based specialized models. InternVL3-8B, Qwen2.5-VL-7B/72B, and Qwen3-VL-8B first perform temporal grounding to predict the frame range, and then apply spatial grounding on frames within the intersection of predicted $b_t$ and ground-truth $b_t^{*}$ for evaluation.

\textbf{Zero-shot.} The object-centric visual prompting paradigm outperforms the two-stage evaluation across InternVL3-8B, Qwen2.5-VL-7B, Qwen2.5-VL-72B and Qwen3-VL-8B, achieving m\_vIoU scores of 15.4\%, 19.5\%, 27.3\%, and 35.8\% on HCSTVG-v2, respectively. The improvement can be attributed to the ability of our paradigm to leverage information from the entire video sequence when generating spatial predictions. However, temporal performance slightly declines for Qwen2.5-VL models due to occlusion of fine-grained details and distributional shifts introduced by visual prompts. Notably, Qwen3-VL-8B exhibits a significant performance boost under our visual prompting paradigm, as the baseline model often fails to detect any target when only static image inputs are provided for dynamically described scenes.

\textbf{Fine-tuning.} Reinforcement learning yields substantial gains in both temporal and spatial performance, establishing new state-of-the-art results on HCSTVG-v1, HCSTVG-v2, and ST-Align. On HCSTVG-v2, compared with the strongest SFT-trained VLM model SpaceVLLM, STVG-R1 achieves absolute improvements of 4.0\%, 6.2\%, 10.9\% and 14.1\% across four evaluation metrics. As shown in Table~\ref{table:STalign}, STVG-R1 also surpasses the strongest ST-Align model LLaVA-ST by +0.6\% on m\_vIoU. These spatial gains highlight the effectiveness of our object-centric visual prompting paradigm in enforcing consistent object-level predictions, while reinforcement learning further enhances reasoning ability, leading to more coherent spatial–temporal video grounding.

\subsection{Evaluation Results on Video Spatial Grounding}
\setlength{\tabcolsep}{2pt}
\begin{wraptable}{r}{0.54\textwidth}
\centering
\small
\renewcommand{\arraystretch}{0.8}
\vspace{-10pt}
\caption{Performance comparison with state-of-the-art models on MeViS (\%). The results of TrackGPT are generated by VISA.}
\begin{tabular}{l|ccc}
\toprule
Models & $J$ & $F$ & $J\&F$ \\\midrule
URVOS \citep{seo2020urvos} & 25.7 & 29.9 & 27.8 \\
MTTR \citep{botach2022end} & 28.8 & 31.2 & 30.0 \\
ReferFormer \citep{wu2022language} & 29.8 & 32.2 & 31.0 \\
LMPM \citep{ding2023mevis} & 34.2 & 40.2 & 37.2 \\
\midrule
LISA \citep{lai2024lisa} & 35.1 & 39.4 & 37.2 \\
TrackGPT \citep{stroh2024trackgpt} & 37.6 & 42.6 & 40.1 \\
VISA \citep{yan2024visa} & 40.7 & 46.3 & 43.5 \\
VideoGlaMM \citep{munasinghe2025videoglamm} & \underline{42.1} & \underline{48.2} & \underline{45.2} \\ \midrule
\rowcolor{gray!15} 
STVG-R1 & \textbf{44.7} & \textbf{50.0} & \textbf{47.3} \\
    \bottomrule
\end{tabular}
\vspace{-5pt}
\label{table:mevis}
\end{wraptable}
\setlength{\tabcolsep}{6pt}
Since vIoU in STVG is inevitably affected by temporal prediction quality, we further evaluate video spatial grounding to isolate spatial capability. As shown in Table~\ref{table:STalign}, the proposed object-centric visual prompting paradigm achieves a notable zero-shot gain of 11.1\% on m\_vIoU on ST-Align video spatial grounding. And after RL, STVG-R1 surpasses the second-best model LLaVA-ST by 13.1\% on m\_vIoU. More importantly, Table~\ref{table:mevis} reports the results on the multi-object referring video object segmentation task. STVG-R1 sets a new state-of-the-art of 47.3\% on $\mathcal{J}\&\mathcal{F}$ on MeViS, despite being trained only on single-object STVG data. This demonstrates the strong generalization ability of our visual prompting paradigm, where the simplified identifier-based formulation facilitates transfer to more complex multi-object scenarios.

\subsection{Evaluation Results on Video Temporal Grounding}
We further evaluate STVG-R1 on out-of-distribution video temporal grounding benchmarks. As shown in Table~\ref{table:videotg}, STVG-R1 achieves the best zero-shot performance on Charades-STA, surpassing the second-best model LLaVA-ST by +7.7\% at tIoU@0.5. Although slightly below the task-specific Time-R1, STVG-R1 achieves competitive results on TVGBench, highlighting strong generalization.
\vspace{-10pt}

\begin{table*}[htbp]
\centering
\small
\renewcommand{\arraystretch}{0.8}
\caption{Performance comparison with state-of-the-art models on Charades-STA and TVGBench (\%). The results marked with $*$ represent models training on corresponding dataset, while others indicate zero-shot settings.}
\begin{tabular}{l|cc|cc}
\toprule
\multirow{2}{*}{Models} & \multicolumn{2}{c|}{Charades-STA} & \multicolumn{2}{c}{TVGBench} \\
\cmidrule(lr){2-3} \cmidrule(lr){4-5}
& tIoU@0.3 & tIoU@0.5 & tIoU@0.3 & tIoU@0.5 \\ \midrule
TimeSuite \citep{zeng2024timesuite} & 69.9 & 48.7 & 31.1 & 18.0 \\
TRACE \citep{guo2024trace} & - & 40.3 & 37.0 & 25.5 \\
LLaVA-ST \citep{li2025llava} & \underline{63.1} & \underline{44.8} & - & - \\
Time-R1 \citep{wang2025time} & 78.1$^{*}$ & 60.8$^{*}$ & \underline{41.8} & \textbf{29.4} \\
\rowcolor{gray!15} \midrule
STVG-R1 & \textbf{73.2} & \textbf{52.5} & \textbf{42.5} & \underline{27.4} \\
\bottomrule
\end{tabular}
\label{table:videotg}
\end{table*}

\vspace{-10pt}
\subsection{Ablation}
\label{sec:ablation}
\setlength{\tabcolsep}{2pt}
\begin{wraptable}{r}{0.54\textwidth}
\centering
\small
\renewcommand{\arraystretch}{0.9}
\vspace{-22pt}
\caption{Ablation study on visual prompt designs on HCSTVG-v1 with zero-shot Qwen2.5-VL-7B. U-Letters denotes uppercase letters, L-Letters denotes lowercase letters, and Mix refers to a combination of numbers and uppercase letters.}
\begin{tabular}{ll|cccc}
\toprule
Size & Type & m\_tIoU & m\_vIoU & vIoU@0.3 & vIoU@0.5 \\
\midrule
10 & Number& 38.1 & 24.6 & 39.4 & 12.0 \\
\rowcolor{gray!15}
20 & Number & 38.0 & \textbf{24.9} & \textbf{40.6} & 12.2 \\
30 & Number & 37.5 & 24.1 & 38.9 & 11.8 \\
40 & Number & 37.4 & 23.2 & 36.4 & 11.6 \\
\midrule
20 & U-Letters & \textbf{39.0} & 24.4 & 38.0 & \textbf{12.6} \\
20 & L-Letters & 38.7 & 24.0 & 37.4 & 12.1 \\
20 & Mix & 38.7 & 15.7 & 20.0 & 5.7 \\
    \bottomrule
\end{tabular}
\label{table:markerdesign1}
\vspace{-10pt}
\end{wraptable}
\setlength{\tabcolsep}{6pt}
\textbf{Ablation on visual prompt design.} Following prior work \citep{cai2024vip, shtedritski2023does}, red-colored visual prompts consistently yield superior performance in identifier recognition. As shown in Table~\ref{table:markerdesign1}, varying the \textbf{font size} leads to only moderate performance differences. Smaller prompts slightly improve temporal grounding, likely due to reduced visual occlusion. Overall, performance remains relatively stable across sizes, and a medium size provides a balanced trade-off between visibility and visual interference. Regarding \textbf{prompt types}, both letters and numbers achieve comparable performance. Letters yield slightly stronger temporal grounding, possibly because they are single-character tokens, whereas multi-digit numbers occupy more space. In contrast, numbers provide marginally better spatial accuracy, likely due to their more consistent visual structure. A mixed design mapping larger numbers to letters does not provide additional benefits, suggesting that simple and consistent prompts are sufficient. Based on these observations, red-colored numeric prompts with font size 20 are adopted as the default configuration for all experiments. 

\setlength{\tabcolsep}{2pt}
\begin{wraptable}{r}{0.46\textwidth}
\centering
\small
\renewcommand{\arraystretch}{0.9}
\vspace{-10pt}
\caption{Experimental results of mask filtering thresholds on HCSTVG-v1. Values before `/' denote the upper bound, and those after `/' are zero-shot results with Qwen2.5-VL-7B.}
\begin{tabular}{l|cccc}
\toprule
$\theta$ & m\_tIoU & m\_vIoU & vIoU@0.3 & vIoU@0.5 \\
\midrule
0 & 100.0/37.5 & 69.1/23.3 & 97.2/37.4 & 89.9/11.9 \\
1/4 & 100.0/38.0 & 68.3/24.6 & 95.8/40.5 & 88.5/11.6 \\
\rowcolor{gray!15}
1/3 & 100.0/38.0 & 67.7/24.9 & 94.6/40.6 & 87.6/12.1 \\
1/2 & 100.0/38.1 & 65.1/24.1 & 89.9/38.2 & 83.5/11.4 \\
    \bottomrule
\end{tabular}
\vspace{-10pt}
\label{table:threshold}
\end{wraptable}
\setlength{\tabcolsep}{6pt}
\textbf{Ablation on mask filtering thresholds.} To reduce visual clutter from dense prompts, we apply mask filtering that removes small instances whose size falls below a fraction of the maximum mask within each category per frame. As shown in Table~\ref{table:threshold}, higher thresholds $\theta$ degrade \textbf{data quality}, with m\_vIoU dropping to 65.1\% at 1/2 on HCSTVG-v1. In contrast, \textbf{zero-shot} evaluation with Qwen2.5-VL-7B shows that moderate filtering improves spatial grounding while maintaining temporal accuracy. A threshold of 1/3 achieves the best trade-off between data quality and zero-shot performance, indicating that many small-scale objects are not semantically salient and may even introduce noise.

\textbf{Ablation on our modules.}
We further conduct ablation studies to explore the individual contributions of object-centric visual prompting paradigm and reinforcement learning. Without visual prompts, GRPO follows the zero-shot two-stage evaluation and is optimized only with the temporal reward. As shown in Table~\ref{table:ablationhcstvg} and Table~\ref{table:ablationstalign}, visual prompts primarily enhance spatial localization, while GRPO substantially improves temporal accuracy. VisualPrompt-SFT improves all metrics but slightly reduces temporal grounding on ST-Align, where reasoning is critical. The combination of visual prompt and GRPO yields the most consistent gains, achieving state-of-the-art performance.

\setlength{\tabcolsep}{2pt}
\begin{table*}[htbp]
\centering
\renewcommand{\arraystretch}{0.8}
\vspace{-15pt}
\caption{Ablation study with different modules on HCSTVG-v1 and HCSTVG-v2.}
\resizebox{\textwidth}{!}{
\begin{tabular}{l|cccc|cccc}
\toprule
\multirow{2}{*}{Models} & \multicolumn{4}{c|}{HCSTVG-v1} & \multicolumn{4}{c}{HCSTVG-v2} \\
\cmidrule(lr){2-5} \cmidrule(lr){6-9}
& m\_tIoU & m\_vIoU & vIoU@0.3 & vIoU@0.5 & m\_tIoU & m\_vIoU & vIoU@0.3 & vIoU@0.5 \\
\midrule
Qwen2.5-VL-7B & 40.3 & 19.7 & 28.2 & 7.9 & 45.1 & 19.3 & 26.0 & 8.2 \\
\rowcolor{gray!15}
\rowcolor{gray!15}
\hspace*{2em}+\textit{VisualPrompt} & 38.7 & 24.8 & 40.7 & 13.4 & 44.7 & 19.2 & 28.7 & 10.9 \\
\rowcolor{gray!15}
\hspace*{2em}+\textit{GRPO} & \textbf{57.5} & 24.7 & 37.8 & 17.6 & 61.4 & 24.2 & 37.7 & 15.4 \\
\rowcolor{gray!15}
\hspace*{2em}+\textit{VisualPrompt-SFT} & 50.9 & 34.3 & 60.2 & 28.0 & 54.4 & 36.5 & 60.8 & 31.3 \\
\rowcolor{gray!15}
\hspace*{2em}+\textit{VisualPrompt-GRPO} & 56.9 & \textbf{39.1} & \textbf{66.7} & \textbf{38.6} & \textbf{62.0} & \textbf{40.2} & \textbf{67.8} & \textbf{38.8} \\
\bottomrule
\end{tabular}
}
\label{table:ablationhcstvg}
\end{table*}
\setlength{\tabcolsep}{6pt}

\setlength{\tabcolsep}{5pt}
\begin{table*}[htbp]
\centering
\renewcommand{\arraystretch}{0.8}
\vspace{-20pt}
\caption{Ablation study with different modules on ST-Align.}
\resizebox{\textwidth}{!}{
\begin{tabular}{l|cccc|cccc}
\toprule
\multirow{2}{*}{Models} & \multicolumn{4}{c|}{Spatial-Temporal Video Grounding} & \multicolumn{3}{c}{Spatial Video Grounding}\\
\cmidrule(lr){2-5} \cmidrule(lr){6-8}
& tIoU@0.5 & m\_tIoU & vIoU@0.5 & m\_vIoU & vIoU@0.3 & vIoU@0.5 & m\_vIoU \\ \midrule
Qwen2.5-VL-7B & 35.2 & 37.4 & 17.1 & 14.3 & 44.6 & 39.5 & 35.5 \\
\rowcolor{gray!15}
\hspace*{2em}+\textit{VisualPrompt} & 36.0 & 38.3 & 21.5 & 19.5 & 57.7 & 51.4 & 46.6 \\
\rowcolor{gray!15}
\hspace*{2em}+\textit{GRPO} & 41.8 & 43.7 & 17.3 & 20.0 & - & - & - \\
\rowcolor{gray!15}
\hspace*{2em}+\textit{VisualPrompt-SFT} & 34.6 & 36.6 & 21.5 & 19.3 & 58.6 & 52.3 & 47.2 \\
\rowcolor{gray!15}
\hspace*{2em}+\textit{VisualPrompt-GRPO} & \textbf{43.6} & \textbf{45.1} & \textbf{25.9} & \textbf{23.4} & \textbf{60.3} & \textbf{53.9} & \textbf{48.6} \\
\bottomrule
\end{tabular}
}
\label{table:ablationstalign}
\end{table*}
\setlength{\tabcolsep}{6pt}

\textbf{Ablation on reward design.} In STVG-R1, the temporal reward is directly derived from the evaluation metric, while the spatial reward adopts a sparse 0/1 formulation that provides credit only when the predicted instance ID matches the ground truth. To examine the role of the spatial component, we tested two alternatives: (1) a coupled spatio-temporal reward inspired by vIoU, defined as $R(o)=r_t(o)+r_s(o),r_t(o)+r_f(o)$, and (2) a continuous spatial reward computed as the average per-frame IoU over the temporal intersection, $r_s=\frac{1}{|\mathcal{T}_{\cap}|}\sum_{t\in\mathcal{T}_{\cap}}\mathrm{IoU}(\hat{B}_t,B_t^\ast)$. Neither variant improves performance. On HCSTVG-v1, the coupled reward reduces m\_vIoU from 39.1\% to 38.3\%, while the continuous spatial reward yields 38.6\%. This suggests that the sparse spatial reward better aligns with the objective of selecting a single correct instance, avoiding additional noise.

\subsection{Impact of Visual Prompt Occlusion}
A potential concern is that overlaying numeric identifiers may introduce visual pollution, particularly for OCR-related tasks. To assess this effect, we evaluate Qwen2.5-VL-7B with and without visual prompts on the MME-VideoOCR~\cite{shi2025mme} benchmark, which covers ten OCR-centric tasks. As shown in Table~\ref{tab:ocr_occlusion}, performance differences remain consistently small across all tasks. Notably, tasks that require fine-grained character recognition, such as TR, show a small performance drop from 69.6\% to 69.3\%. In contrast, higher-level tasks such as VTQA and TG improve from 76.4\% to 77.4\% and from 63.2\% to 65.3\%, suggesting that the added markers can help guide attention toward relevant regions. These results indicate that the visual prompts introduce negligible interference.

\setlength{\tabcolsep}{4pt}
\begin{table}[!h]
\centering
\caption{Evaluation results on MME-VideoOCR. `TR' denotes Text Recognition, `VTQA' Visual Text QA, `TG' Text Grounding, `AR' Attribute Recognition, `CDT' Change Detection \& Tracking, `STP' Special Text Parsing, `CFTU' Cross-Frame Text Understanding, `TBR' Text-Based Reasoning, `TBVU' Text-Based Video Understanding, and `RVT' Robust Video Testing.}
\label{tab:ocr_occlusion}
\begin{tabular}{lccccccccccc}
\toprule
Method & TR & VTQA & TG & AR & CDT & STP & CFTU & TBR & TBVU & RVT & Total \\
\midrule
Qwen2.5-VL-7B & \textbf{69.6} & 76.4 & 63.2 & \textbf{69.8} & 45.7 & \textbf{64.4} & \textbf{22.7} & 54.7 & \textbf{38.2} & \textbf{79.0} & \textbf{59.4} \\
+ VisualPrompt & 69.3 & \textbf{77.4} & \textbf{65.3} & 68.5 & \textbf{46.2} & 64.1 & 22.7 & \textbf{54.8} & 38.2 & 78.8 & 58.9 \\
\bottomrule
\end{tabular}
\end{table}
\setlength{\tabcolsep}{6pt}

\subsection{Visualization}
\begin{figure}[htbp]
\begin{center}
\includegraphics[width=1\textwidth]{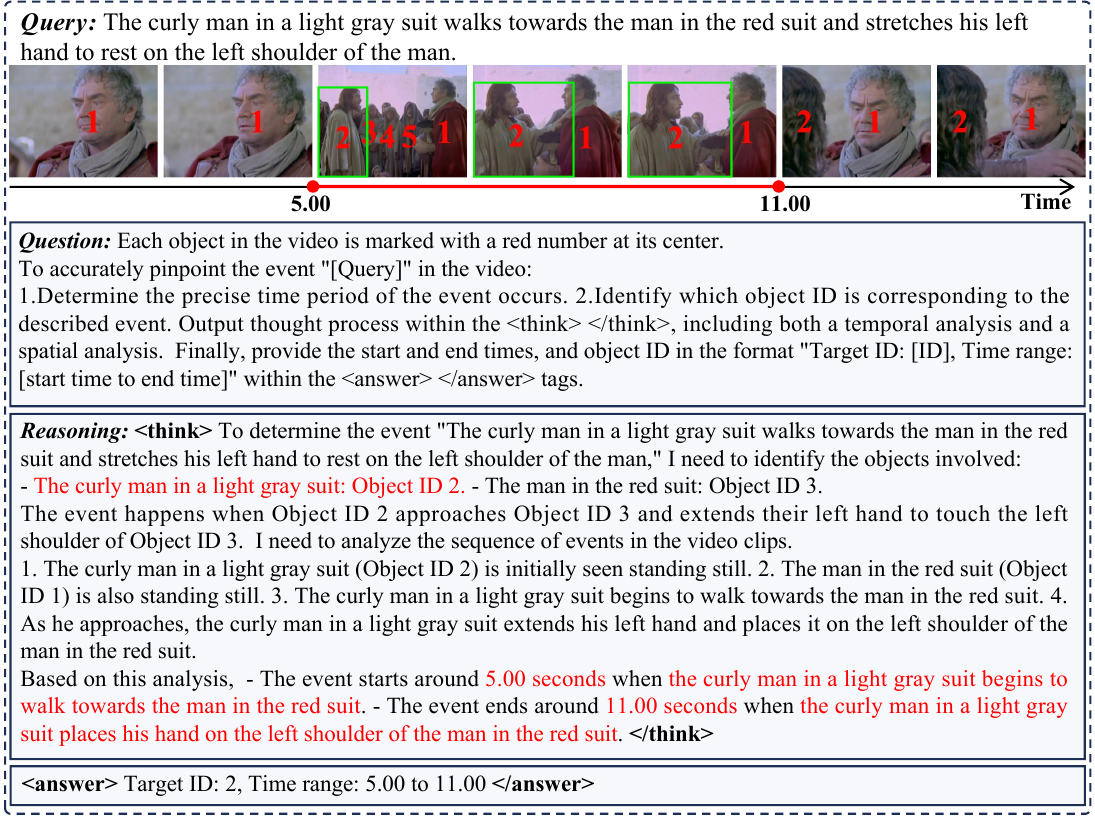}
\end{center}
\setlength{\abovecaptionskip}{0pt}
\caption{Case study of STVG-R1 on the spatial-temporal video grounding task.}
\label{fig:case-stvg}
\end{figure}

In Figure~\ref{fig:case-stvg}, we present a case of spatial–temporal video grounding with our object-centric prompting paradigm. The model first identifies the object IDs relevant to the query and then determines the temporal boundaries. During the reasoning process, complex descriptive appearance expressions are compactly mapped into object IDs, facilitating precise instance-level reasoning and grounding.

\section{Conclusion}
This work addresses challenges of coordinates visual–textual misalignment and instance prediction inconsistency across videos. We propose an object-centric visual prompting paradigm that reformulates per-frame coordinate prediction into a compact and interpretable instance-level identification problem. We further introduce STVG-R1, a reinforcement learning framework optimized with task-driven rewards. Experiments across six benchmarks demonstrate the effectiveness of compact visual prompts and reinforcement learning in enhancing reasoning consistency and generalization. Future work will extend our detector-based framework from natural images to broader visual domains.

\section{REPRODUCIBILITY STATEMENT}
We make every effort to ensure the reproducibility of our work. Detailed descriptions of the model architecture, training pipeline, training datasets, and reward design for STVG-R1 are provided in Section 3. Implementation details are reported in Section 4.1. The design of visual prompts and filtering thresholds is described in Section 4.5, and the prompts used for training and evaluation across different tasks are presented in Section A.1.

\bibliography{iclr2026_conference}

@inproceedings{gu2024context,
  title={Context-guided spatio-temporal video grounding},
  author={Gu, Xin and Fan, Heng and Huang, Yan and Luo, Tiejian and Zhang, Libo},
  booktitle={Proceedings of the IEEE/CVF Conference on Computer Vision and Pattern Recognition},
  pages={18330--18339},
  year={2024}
}

@article{gu2025knowing,
  title={Knowing your target: Target-aware transformer makes better spatio-temporal video grounding},
  author={Gu, Xin and Shen, Yaojie and Luo, Chenxi and Luo, Tiejian and Huang, Yan and Lin, Yuewei and Fan, Heng and Zhang, Libo},
  journal={arXiv preprint arXiv:2502.11168},
  year={2025}
}

@inproceedings{carreira2017quo,
  title={Quo vadis, action recognition? a new model and the kinetics dataset},
  author={Carreira, Joao and Zisserman, Andrew},
  booktitle={proceedings of the IEEE Conference on Computer Vision and Pattern Recognition},
  pages={6299--6308},
  year={2017}
}

@inproceedings{radford2021learning,
  title={Learning transferable visual models from natural language supervision},
  author={Radford, Alec and Kim, Jong Wook and Hallacy, Chris and Ramesh, Aditya and Goh, Gabriel and Agarwal, Sandhini and Sastry, Girish and Askell, Amanda and Mishkin, Pamela and Clark, Jack and others},
  booktitle={International conference on machine learning},
  pages={8748--8763},
  year={2021},
  organization={PmLR}
}

@inproceedings{lin2022stvgformer,
  title={Stvgformer: Spatio-temporal video grounding with static-dynamic cross-modal understanding},
  author={Lin, Zihang and Tan, Chaolei and Hu, Jian-Fang and Jin, Zhi and Ye, Tiancai and Zheng, Wei-Shi},
  booktitle={Proceedings of the 4th on Person in Context Workshop},
  pages={1--5},
  year={2022}
}

@article{li2024llava,
  title={Llava-onevision: Easy visual task transfer},
  author={Li, Bo and Zhang, Yuanhan and Guo, Dong and Zhang, Renrui and Li, Feng and Zhang, Hao and Zhang, Kaichen and Zhang, Peiyuan and Li, Yanwei and Liu, Ziwei and others},
  journal={arXiv preprint arXiv:2408.03326},
  year={2024}
}

@inproceedings{wang2024internvideo2,
  title={Internvideo2: Scaling foundation models for multimodal video understanding},
  author={Wang, Yi and Li, Kunchang and Li, Xinhao and Yu, Jiashuo and He, Yinan and Chen, Guo and Pei, Baoqi and Zheng, Rongkun and Wang, Zun and Shi, Yansong and others},
  booktitle={European Conference on Computer Vision},
  pages={396--416},
  year={2024},
  organization={Springer}
}

@article{bai2025qwen2,
  title={Qwen2. 5-vl technical report},
  author={Bai, Shuai and Chen, Keqin and Liu, Xuejing and Wang, Jialin and Ge, Wenbin and Song, Sibo and Dang, Kai and Wang, Peng and Wang, Shijie and Tang, Jun and others},
  journal={arXiv preprint arXiv:2502.13923},
  year={2025}
}

@article{abdin2024phi,
  title={Phi-4 technical report},
  author={Abdin, Marah and Aneja, Jyoti and Behl, Harkirat and Bubeck, S{\'e}bastien and Eldan, Ronen and Gunasekar, Suriya and Harrison, Michael and Hewett, Russell J and Javaheripi, Mojan and Kauffmann, Piero and others},
  journal={arXiv preprint arXiv:2412.08905},
  year={2024}
}

@article{zhang2024video,
  title={Video instruction tuning with synthetic data},
  author={Zhang, Yuanhan and Wu, Jinming and Li, Wei and Li, Bo and Ma, Zejun and Liu, Ziwei and Li, Chunyuan},
  journal={arXiv preprint arXiv:2410.02713},
  year={2024}
}

@inproceedings{li2025llava,
  title={Llava-st: A multimodal large language model for fine-grained spatial-temporal understanding},
  author={Li, Hongyu and Chen, Jinyu and Wei, Ziyu and Huang, Shaofei and Hui, Tianrui and Gao, Jialin and Wei, Xiaoming and Liu, Si},
  booktitle={Proceedings of the Computer Vision and Pattern Recognition Conference},
  pages={8592--8603},
  year={2025}
}

@article{yuan2025sa2va,
  title={Sa2va: Marrying sam2 with llava for dense grounded understanding of images and videos},
  author={Yuan, Haobo and Li, Xiangtai and Zhang, Tao and Huang, Zilong and Xu, Shilin and Ji, Shunping and Tong, Yunhai and Qi, Lu and Feng, Jiashi and Yang, Ming-Hsuan},
  journal={arXiv preprint arXiv:2501.04001},
  year={2025}
}

@inproceedings{munasinghe2025videoglamm,
  title={Videoglamm: A large multimodal model for pixel-level visual grounding in videos},
  author={Munasinghe, Shehan and Gani, Hanan and Zhu, Wenqi and Cao, Jiale and Xing, Eric and Khan, Fahad Shahbaz and Khan, Salman},
  booktitle={Proceedings of the Computer Vision and Pattern Recognition Conference},
  pages={19036--19046},
  year={2025}
}

@article{sun2025sama,
  title={SAMA: Towards Multi-Turn Referential Grounded Video Chat with Large Language Models},
  author={Sun, Ye and Zhang, Hao and Ding, Henghui and Zhang, Tiehua and Ma, Xingjun and Jiang, Yu-Gang},
  journal={arXiv preprint arXiv:2505.18812},
  year={2025}
}

@article{wang2025spacevllm,
  title={SpaceVLLM: Endowing Multimodal Large Language Model with Spatio-Temporal Video Grounding Capability},
  author={Wang, Jiankang and Zhang, Zhihan and Liu, Zhihang and Li, Yang and Ge, Jiannan and Xie, Hongtao and Zhang, Yongdong},
  journal={arXiv preprint arXiv:2503.13983},
  year={2025}
}

@inproceedings{devlin2019bert,
  title={Bert: Pre-training of deep bidirectional transformers for language understanding},
  author={Devlin, Jacob and Chang, Ming-Wei and Lee, Kenton and Toutanova, Kristina},
  booktitle={Proceedings of the 2019 conference of the North American chapter of the association for computational linguistics: human language technologies, volume 1 (long and short papers)},
  pages={4171--4186},
  year={2019}
}

@article{tang2021human,
  title={Human-centric spatio-temporal video grounding with visual transformers},
  author={Tang, Zongheng and Liao, Yue and Liu, Si and Li, Guanbin and Jin, Xiaojie and Jiang, Hongxu and Yu, Qian and Xu, Dong},
  journal={IEEE Transactions on Circuits and Systems for Video Technology},
  volume={32},
  number={12},
  pages={8238--8249},
  year={2021},
  publisher={IEEE}
}

@inproceedings{zhang2020does,
  title={Where does it exist: Spatio-temporal video grounding for multi-form sentences},
  author={Zhang, Zhu and Zhao, Zhou and Zhao, Yang and Wang, Qi and Liu, Huasheng and Gao, Lianli},
  booktitle={Proceedings of the IEEE/CVF Conference on Computer Vision and Pattern Recognition},
  pages={10668--10677},
  year={2020}
}

@article{jaech2024openai,
  title={Openai o1 system card},
  author={Jaech, Aaron and Kalai, Adam and Lerer, Adam and Richardson, Adam and El-Kishky, Ahmed and Low, Aiden and Helyar, Alec and Madry, Aleksander and Beutel, Alex and Carney, Alex and others},
  journal={arXiv preprint arXiv:2412.16720},
  year={2024}
}

@article{liu2025visual,
  title={Visual-rft: Visual reinforcement fine-tuning},
  author={Liu, Ziyu and Sun, Zeyi and Zang, Yuhang and Dong, Xiaoyi and Cao, Yuhang and Duan, Haodong and Lin, Dahua and Wang, Jiaqi},
  journal={arXiv preprint arXiv:2503.01785},
  year={2025}
}

@article{feng2025video,
  title={Video-r1: Reinforcing video reasoning in mllms},
  author={Feng, Kaituo and Gong, Kaixiong and Li, Bohao and Guo, Zonghao and Wang, Yibing and Peng, Tianshuo and Wu, Junfei and Zhang, Xiaoying and Wang, Benyou and Yue, Xiangyu},
  journal={arXiv preprint arXiv:2503.21776},
  year={2025}
}

@article{zhang2025r1,
  title={R1-vl: Learning to reason with multimodal large language models via step-wise group relative policy optimization},
  author={Zhang, Jingyi and Huang, Jiaxing and Yao, Huanjin and Liu, Shunyu and Zhang, Xikun and Lu, Shijian and Tao, Dacheng},
  journal={arXiv preprint arXiv:2503.12937},
  year={2025}
}

@article{shen2025vlm,
  title={Vlm-r1: A stable and generalizable r1-style large vision-language model},
  author={Shen, Haozhan and Liu, Peng and Li, Jingcheng and Fang, Chunxin and Ma, Yibo and Liao, Jiajia and Shen, Qiaoli and Zhang, Zilun and Zhao, Kangjia and Zhang, Qianqian and others},
  journal={arXiv preprint arXiv:2504.07615},
  year={2025}
}

@article{guo2025deepseek,
  title={Deepseek-r1: Incentivizing reasoning capability in llms via reinforcement learning},
  author={Guo, Daya and Yang, Dejian and Zhang, Haowei and Song, Junxiao and Zhang, Ruoyu and Xu, Runxin and Zhu, Qihao and Ma, Shirong and Wang, Peiyi and Bi, Xiao and others},
  journal={arXiv preprint arXiv:2501.12948},
  year={2025}
}

@article{wang2025time,
  title={Time-R1: Post-Training Large Vision Language Model for Temporal Video Grounding},
  author={Wang, Ye and Wang, Ziheng and Xu, Boshen and Du, Yang and Lin, Kejun and Xiao, Zihan and Yue, Zihao and Ju, Jianzhong and Zhang, Liang and Yang, Dingyi and others},
  journal={arXiv preprint arXiv:2503.13377},
  year={2025}
}

@article{shao2024deepseekmath,
  title={Deepseekmath: Pushing the limits of mathematical reasoning in open language models},
  author={Shao, Zhihong and Wang, Peiyi and Zhu, Qihao and Xu, Runxin and Song, Junxiao and Bi, Xiao and Zhang, Haowei and Zhang, Mingchuan and Li, YK and Wu, Yang and others},
  journal={arXiv preprint arXiv:2402.03300},
  year={2024}
}

@misc{chen2025r1v,
  author       = {Chen, Liang and Li, Lei and Zhao, Haozhe and Song, Yifan and Vinci},
  title        = {R1-V: Reinforcing Super Generalization Ability in Vision-Language Models with Less Than \$3},
  howpublished = {\url{https://github.com/Deep-Agent/R1-V}},
  note         = {Accessed: 2025-02-02},
  year         = {2025}
}

@article{wang2024qwen2,
  title={Qwen2-vl: Enhancing vision-language model's perception of the world at any resolution},
  author={Wang, Peng and Bai, Shuai and Tan, Sinan and Wang, Shijie and Fan, Zhihao and Bai, Jinze and Chen, Keqin and Liu, Xuejing and Wang, Jialin and Ge, Wenbin and others},
  journal={arXiv preprint arXiv:2409.12191},
  year={2024}
}

@inproceedings{lin2024vila,
  title={Vila: On pre-training for visual language models},
  author={Lin, Ji and Yin, Hongxu and Ping, Wei and Molchanov, Pavlo and Shoeybi, Mohammad and Han, Song},
  booktitle={Proceedings of the IEEE/CVF conference on computer vision and pattern recognition},
  pages={26689--26699},
  year={2024}
}

@inproceedings{liu2024improved,
  title={Improved baselines with visual instruction tuning},
  author={Liu, Haotian and Li, Chunyuan and Li, Yuheng and Lee, Yong Jae},
  booktitle={Proceedings of the IEEE/CVF conference on computer vision and pattern recognition},
  pages={26296--26306},
  year={2024}
}

@article{wang2024cogvlm,
  title={Cogvlm: Visual expert for pretrained language models},
  author={Wang, Weihan and Lv, Qingsong and Yu, Wenmeng and Hong, Wenyi and Qi, Ji and Wang, Yan and Ji, Junhui and Yang, Zhuoyi and Zhao, Lei and XiXuan, Song and others},
  journal={Advances in Neural Information Processing Systems},
  volume={37},
  pages={121475--121499},
  year={2024}
}

@article{chen2024longvila,
  title={Longvila: Scaling long-context visual language models for long videos},
  author={Chen, Yukang and Xue, Fuzhao and Li, Dacheng and Hu, Qinghao and Zhu, Ligeng and Li, Xiuyu and Fang, Yunhao and Tang, Haotian and Yang, Shang and Liu, Zhijian and others},
  journal={arXiv preprint arXiv:2408.10188},
  year={2024}
}

@article{qi2025gpt4scene,
  title={Gpt4scene: Understand 3d scenes from videos with vision-language models},
  author={Qi, Zhangyang and Zhang, Zhixiong and Fang, Ye and Wang, Jiaqi and Zhao, Hengshuang},
  journal={arXiv preprint arXiv:2501.01428},
  year={2025}
}

@inproceedings{shtedritski2023does,
  title={What does clip know about a red circle? visual prompt engineering for vlms},
  author={Shtedritski, Aleksandar and Rupprecht, Christian and Vedaldi, Andrea},
  booktitle={Proceedings of the IEEE/CVF International Conference on Computer Vision},
  pages={11987--11997},
  year={2023}
}

@inproceedings{cai2024vip,
  title={Vip-llava: Making large multimodal models understand arbitrary visual prompts},
  author={Cai, Mu and Liu, Haotian and Mustikovela, Siva Karthik and Meyer, Gregory P and Chai, Yuning and Park, Dennis and Lee, Yong Jae},
  booktitle={Proceedings of the IEEE/CVF Conference on Computer Vision and Pattern Recognition},
  pages={12914--12923},
  year={2024}
}

@article{yang2024set,
  title={Set-of-mark prompting unleashes extraordinary visual grounding in gpt-4v, 2023},
  author={Yang, Jianwei and Zhang, Hao and Li, Feng and Zou, Xueyan and Li, Chunyuan and Gao, Jianfeng},
  journal={URL https://arxiv. org/abs/2310.11441},
  volume={3},
  year={2024}
}

@article{tian2025yolov12,
  title={YOLOv12: Attention-Centric Real-Time Object Detectors},
  author={Tian, Yunjie and Ye, Qixiang and Doermann, David},
  journal={arXiv preprint arXiv:2502.12524},
  year={2025}
}

@article{ravi2024sam2,
  title={SAM 2: Segment Anything in Images and Videos},
  author={Ravi, Nikhila and Gabeur, Valentin and Hu, Yuan-Ting and Hu, Ronghang and Ryali, Chaitanya and Ma, Tengyu and Khedr, Haitham and R{\"a}dle, Roman and Rolland, Chloe and Gustafson, Laura and Mintun, Eric and Pan, Junting and Alwala, Kalyan Vasudev and Carion, Nicolas and Wu, Chao-Yuan and Girshick, Ross and Doll{\'a}r, Piotr and Feichtenhofer, Christoph},
  journal={arXiv preprint arXiv:2408.00714},
  url={https://arxiv.org/abs/2408.00714},
  year={2024}
}

@inproceedings{lin2023collaborative,
  title={Collaborative static and dynamic vision-language streams for spatio-temporal video grounding},
  author={Lin, Zihang and Tan, Chaolei and Hu, Jian-Fang and Jin, Zhi and Ye, Tiancai and Zheng, Wei-Shi},
  booktitle={Proceedings of the IEEE/CVF Conference on Computer Vision and Pattern Recognition},
  pages={23100--23109},
  year={2023}
}

@inproceedings{yang2022tubedetr,
  title={Tubedetr: Spatio-temporal video grounding with transformers},
  author={Yang, Antoine and Miech, Antoine and Sivic, Josef and Laptev, Ivan and Schmid, Cordelia},
  booktitle={Proceedings of the IEEE/CVF Conference on Computer Vision and Pattern Recognition},
  pages={16442--16453},
  year={2022}
}

@article{loshchilov2017decoupled,
  title={Decoupled weight decay regularization},
  author={Loshchilov, Ilya and Hutter, Frank},
  journal={arXiv preprint arXiv:1711.05101},
  year={2017}
}

@article{zeng2024timesuite,
  title={Timesuite: Improving mllms for long video understanding via grounded tuning},
  author={Zeng, Xiangyu and Li, Kunchang and Wang, Chenting and Li, Xinhao and Jiang, Tianxiang and Yan, Ziang and Li, Songze and Shi, Yansong and Yue, Zhengrong and Wang, Yi and others},
  journal={arXiv preprint arXiv:2410.19702},
  year={2024}
}

@article{guo2024trace,
  title={Trace: Temporal grounding video llm via causal event modeling},
  author={Guo, Yongxin and Liu, Jingyu and Li, Mingda and Liu, Qingbin and Chen, Xi and Tang, Xiaoying},
  journal={arXiv preprint arXiv:2410.05643},
  year={2024}
}

@inproceedings{gao2017tall,
  title={Tall: Temporal activity localization via language query},
  author={Gao, Jiyang and Sun, Chen and Yang, Zhenheng and Nevatia, Ram},
  booktitle={Proceedings of the IEEE international conference on computer vision},
  pages={5267--5275},
  year={2017}
}

@inproceedings{ding2023mevis,
  title={MeViS: A large-scale benchmark for video segmentation with motion expressions},
  author={Ding, Henghui and Liu, Chang and He, Shuting and Jiang, Xudong and Loy, Chen Change},
  booktitle={Proceedings of the IEEE/CVF international conference on computer vision},
  pages={2694--2703},
  year={2023}
}

@inproceedings{yan2024visa,
  title={Visa: Reasoning video object segmentation via large language models},
  author={Yan, Cilin and Wang, Haochen and Yan, Shilin and Jiang, Xiaolong and Hu, Yao and Kang, Guoliang and Xie, Weidi and Gavves, Efstratios},
  booktitle={European Conference on Computer Vision},
  pages={98--115},
  year={2024},
  organization={Springer}
}

@article{golomb1966run,
  title={Run-length encodings (corresp.)},
  author={Golomb, Solomon},
  journal={IEEE transactions on information theory},
  volume={12},
  number={3},
  pages={399--401},
  year={1966},
  publisher={IEEE}
}

@inproceedings{yang2025thinking,
  title={Thinking in space: How multimodal large language models see, remember, and recall spaces},
  author={Yang, Jihan and Yang, Shusheng and Gupta, Anjali W and Han, Rilyn and Fei-Fei, Li and Xie, Saining},
  booktitle={Proceedings of the Computer Vision and Pattern Recognition Conference},
  pages={10632--10643},
  year={2025}
}

@inproceedings{fu2025video,
  title={Video-mme: The first-ever comprehensive evaluation benchmark of multi-modal llms in video analysis},
  author={Fu, Chaoyou and Dai, Yuhan and Luo, Yongdong and Li, Lei and Ren, Shuhuai and Zhang, Renrui and Wang, Zihan and Zhou, Chenyu and Shen, Yunhang and Zhang, Mengdan and others},
  booktitle={Proceedings of the Computer Vision and Pattern Recognition Conference},
  pages={24108--24118},
  year={2025}
}

@article{zhu2025internvl3,
  title={Internvl3: Exploring advanced training and test-time recipes for open-source multimodal models},
  author={Zhu, Jinguo and Wang, Weiyun and Chen, Zhe and Liu, Zhaoyang and Ye, Shenglong and Gu, Lixin and Tian, Hao and Duan, Yuchen and Su, Weijie and Shao, Jie and others},
  journal={arXiv preprint arXiv:2504.10479},
  year={2025}
}

@inproceedings{liu2024grounding,
  title={Grounding dino: Marrying dino with grounded pre-training for open-set object detection},
  author={Liu, Shilong and Zeng, Zhaoyang and Ren, Tianhe and Li, Feng and Zhang, Hao and Yang, Jie and Jiang, Qing and Li, Chunyuan and Yang, Jianwei and Su, Hang and others},
  booktitle={European conference on computer vision},
  pages={38--55},
  year={2024},
  organization={Springer}
}

@article{xiao2023florence,
  title={Florence-2: Advancing a unified representation for a variety of vision tasks},
  author={Xiao, Bin and Wu, Haiping and Xu, Weijian and Dai, Xiyang and Hu, Houdong and Lu, Yumao and Zeng, Michael and Liu, Ce and Yuan, Lu},
  journal={arXiv preprint arXiv:2311.06242},
  year={2023}
}

@article{ye2024x,
  title={X-vila: Cross-modality alignment for large language model},
  author={Ye, Hanrong and Huang, De-An and Lu, Yao and Yu, Zhiding and Ping, Wei and Tao, Andrew and Kautz, Jan and Han, Song and Xu, Dan and Molchanov, Pavlo and others},
  journal={arXiv preprint arXiv:2405.19335},
  year={2024}
}

@inproceedings{seo2020urvos,
  title={Urvos: Unified referring video object segmentation network with a large-scale benchmark},
  author={Seo, Seonguk and Lee, Joon-Young and Han, Bohyung},
  booktitle={European conference on computer vision},
  pages={208--223},
  year={2020},
  organization={Springer}
}

@inproceedings{botach2022end,
  title={End-to-end referring video object segmentation with multimodal transformers},
  author={Botach, Adam and Zheltonozhskii, Evgenii and Baskin, Chaim},
  booktitle={Proceedings of the IEEE/CVF Conference on Computer Vision and Pattern Recognition},
  pages={4985--4995},
  year={2022}
}

@inproceedings{wu2022language,
  title={Language as queries for referring video object segmentation},
  author={Wu, Jiannan and Jiang, Yi and Sun, Peize and Yuan, Zehuan and Luo, Ping},
  booktitle={Proceedings of the IEEE/CVF Conference on Computer Vision and Pattern Recognition},
  pages={4974--4984},
  year={2022}
}

@inproceedings{lai2024lisa,
  title={Lisa: Reasoning segmentation via large language model},
  author={Lai, Xin and Tian, Zhuotao and Chen, Yukang and Li, Yanwei and Yuan, Yuhui and Liu, Shu and Jia, Jiaya},
  booktitle={Proceedings of the IEEE/CVF Conference on Computer Vision and Pattern Recognition},
  pages={9579--9589},
  year={2024}
}

@article{stroh2024trackgpt,
  title={TrackGPT--A generative pre-trained transformer for cross-domain entity trajectory forecasting},
  author={Stroh, Nicholas},
  journal={arXiv preprint arXiv:2402.00066},
  year={2024}
}

@article{chen2025scaling,
  title={Scaling rl to long videos},
  author={Chen, Yukang and Huang, Wei and Shi, Baifeng and Hu, Qinghao and Ye, Hanrong and Zhu, Ligeng and Liu, Zhijian and Molchanov, Pavlo and Kautz, Jan and Qi, Xiaojuan and others},
  journal={arXiv preprint arXiv:2507.07966},
  year={2025}
}

@article{khan2022grounded,
  title={Grounded video situation recognition},
  author={Khan, Zeeshan and Jawahar, CV and Tapaswi, Makarand},
  journal={Advances in Neural Information Processing Systems},
  volume={35},
  pages={8199--8210},
  year={2022}
}

@inproceedings{grauman2022ego4d,
  title={Ego4d: Around the world in 3,000 hours of egocentric video},
  author={Grauman, Kristen and Westbury, Andrew and Byrne, Eugene and Chavis, Zachary and Furnari, Antonino and Girdhar, Rohit and Hamburger, Jackson and Jiang, Hao and Liu, Miao and Liu, Xingyu and others},
  booktitle={Proceedings of the IEEE/CVF conference on computer vision and pattern recognition},
  pages={18995--19012},
  year={2022}
}

@article{shi2025mme,
  title={Mme-videoocr: Evaluating ocr-based capabilities of multimodal llms in video scenarios},
  author={Shi, Yang and Wang, Huanqian and Xie, Wulin and Zhang, Huanyao and Zhao, Lijie and Zhang, Yi-Fan and Li, Xinfeng and Fu, Chaoyou and Wen, Zhuoer and Liu, Wenting and others},
  journal={arXiv preprint arXiv:2505.21333},
  year={2025}
}

@inproceedings{fan2024videoagent,
  title={Videoagent: A memory-augmented multimodal agent for video understanding},
  author={Fan, Yue and Ma, Xiaojian and Wu, Rujie and Du, Yuntao and Li, Jiaqi and Gao, Zhi and Li, Qing},
  booktitle={European Conference on Computer Vision},
  pages={75--92},
  year={2024},
  organization={Springer}
}

@inproceedings{fan2025embodied,
  title={Embodied videoagent: Persistent memory from egocentric videos and embodied sensors enables dynamic scene understanding},
  author={Fan, Yue and Ma, Xiaojian and Su, Rongpeng and Guo, Jun and Wu, Rujie and Chen, Xi and Li, Qing},
  booktitle={International Conference on Computer Vision},
  year={2025},
}

@article{zhang2025chain,
  title={Chain-of-Focus: Adaptive Visual Search and Zooming for Multimodal Reasoning via RL},
  author={Zhang, Xintong and Gao, Zhi and Zhang, Bofei and Li, Pengxiang and Zhang, Xiaowen and Liu, Yang and Yuan, Tao and Wu, Yuwei and Jia, Yunde and Zhu, Song-Chun and others},
  journal={arXiv preprint arXiv:2505.15436},
  year={2025}
}
\bibliographystyle{iclr2026_conference}

\newpage
\appendix
\section{Appendix}
\subsection{Prompt for Training and Evaluation}
\label{sec:prompt}
\begin{figure}[htbp]
\begin{center}
\includegraphics[width=1\textwidth]{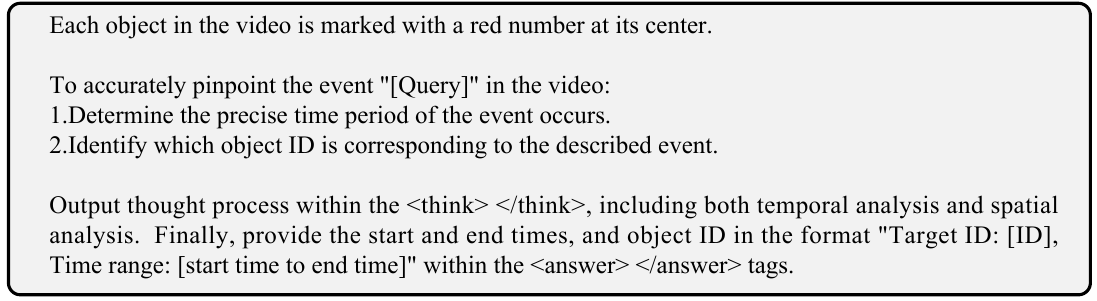}
\end{center}
\setlength{\abovecaptionskip}{0pt}
\caption{Prompt for spatial-temporal video grounding.}
\label{fig:prompt1}
\end{figure}

The prompt for spatial-temporal video grounding is shown in Figure~\ref{fig:prompt1}.

\begin{figure}[htbp]
\begin{center}
\includegraphics[width=1\textwidth]{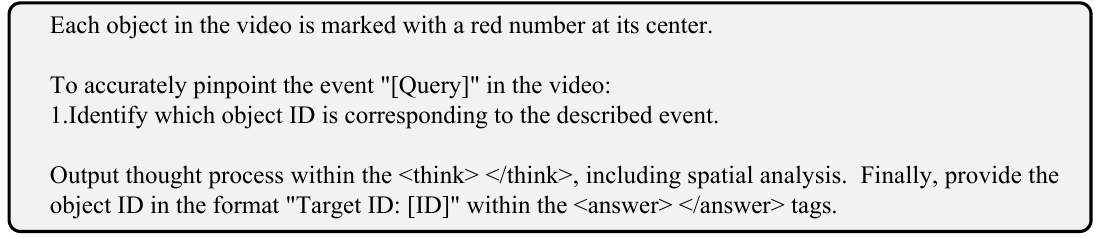}
\end{center}
\setlength{\abovecaptionskip}{0pt}
\caption{Prompt for video spatial grounding and referring video object segmentation.}
\label{fig:prompt2}
\end{figure}

The prompt for video spatial grounding and referring video object segmentation is shown in Figure~\ref{fig:prompt2}.

\begin{figure}[htbp]
\begin{center}
\includegraphics[width=1\textwidth]{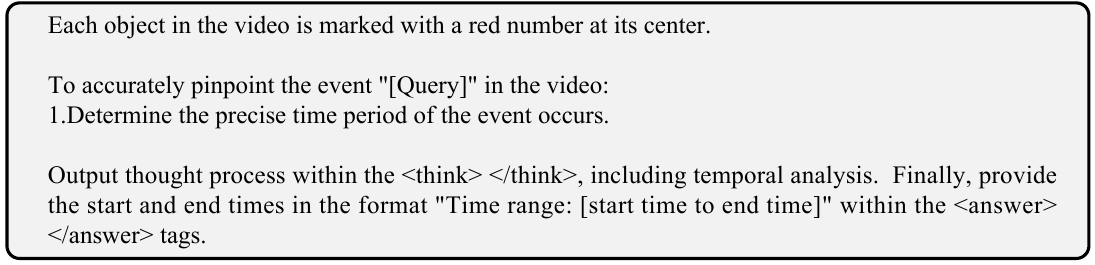}
\end{center}
\setlength{\abovecaptionskip}{0pt}
\caption{Prompt for video temporal grounding.}
\label{fig:prompt3}
\end{figure}

The prompt for video temporal grounding is shown in Figure~\ref{fig:prompt3}.

\subsection{Visualizations of Different Tasks}
\begin{figure}[htbp]
\begin{center}
\includegraphics[width=1\textwidth]{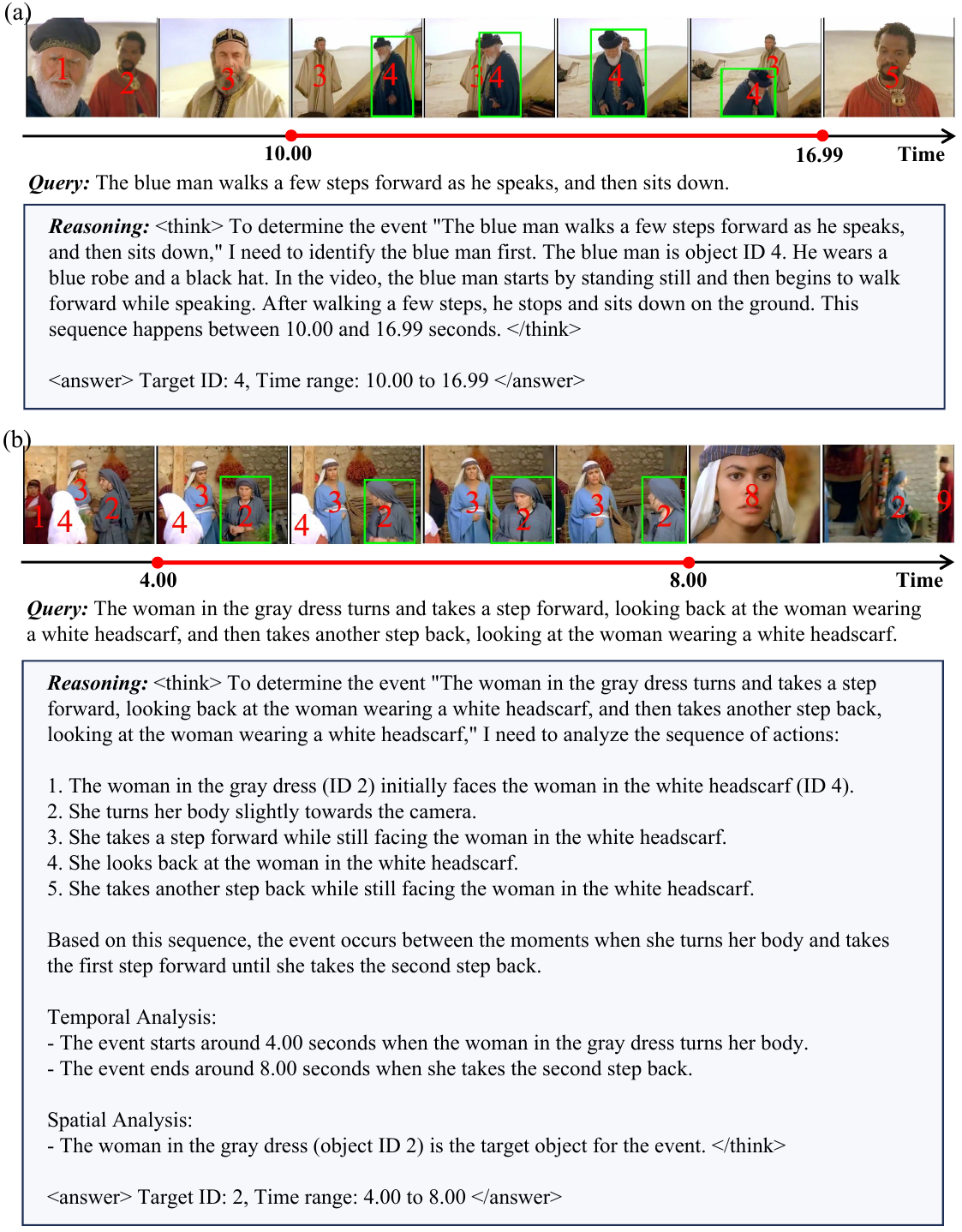}
\end{center}
\setlength{\abovecaptionskip}{0pt}
\caption{Cases for spatial-temporal video grounding.}
\label{fig:vis-stvg}
\end{figure}

\begin{figure}[htbp]
\begin{center}
\includegraphics[width=1\textwidth]{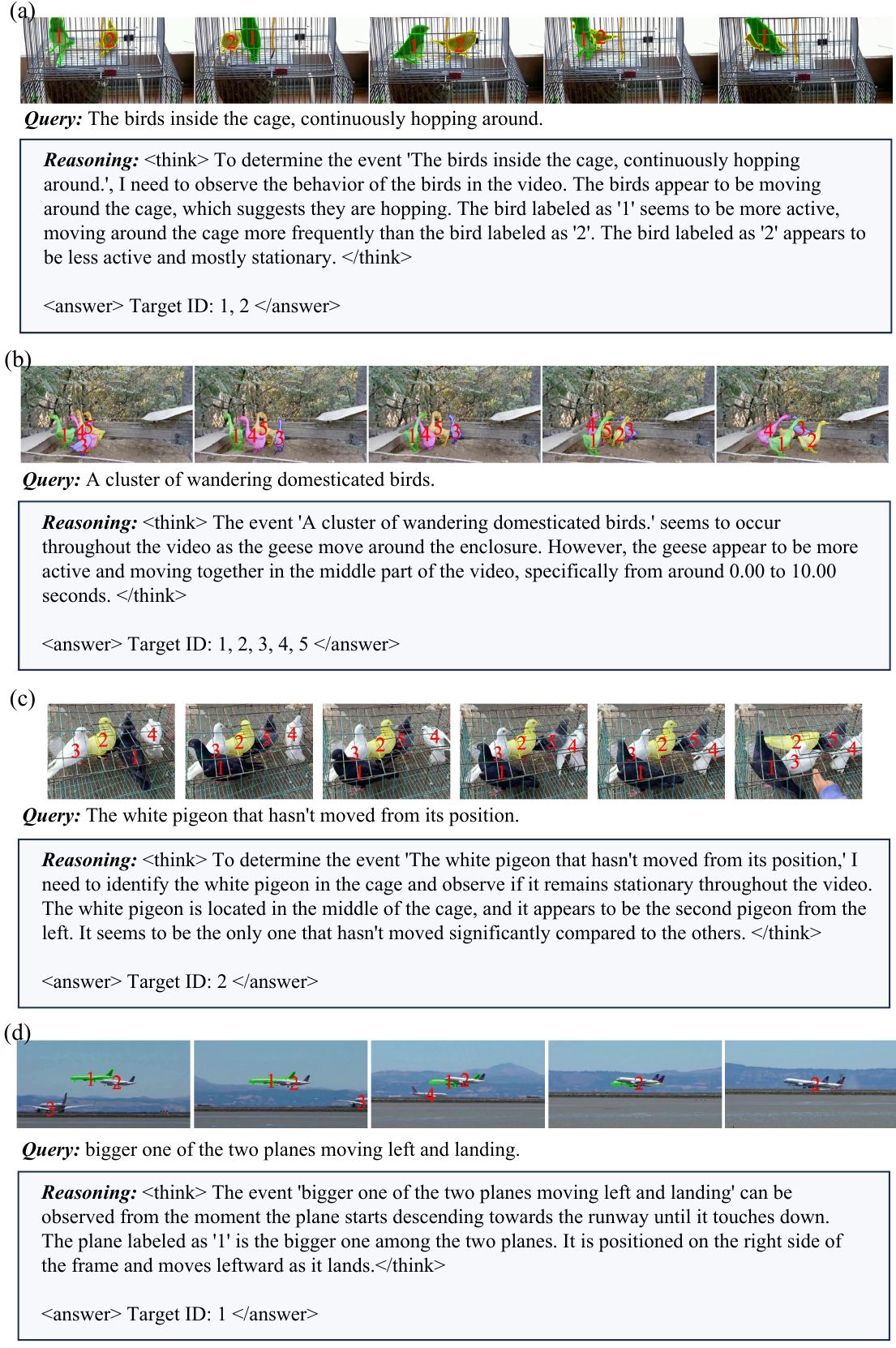}
\end{center}
\setlength{\abovecaptionskip}{0pt}
\caption{Cases for referring video object segmentation.}
\label{fig:vis-refvos}
\end{figure}

We provide more cases of spatial-temporal video grounding and referring video object segmentation tasks, as presented in Figure~\ref{fig:vis-stvg} and Figure~\ref{fig:vis-refvos}. Although some visual prompts are filtered out in certain frames, the corresponding instances remain stored in the mask database, thereby enhancing the final spatial localization capability.

\newpage
\subsection{More Experiments}
Since video temporal grounding task does not inherently require object-centric visual prompts, their addition may even obscure fine-grained video details and slightly hinder temporal accuracy in the zero-shot setting. To further verify this, we compare STVG-R1 with and without visual prompts on Charades-STA and TVGBench. As shown in Table~\ref{table:timep}, visual prompts bring a marginal decrease in performance, indicating that visual prompts are not essential for temporal-only task.

\begin{table*}[htbp]
\centering
\small
\renewcommand{\arraystretch}{0.9}
\caption{Comparison of STVG-R1 with and without visual prompts on temporal grounding benchmarks Charades-STA and TVGBench (\%). Adding visual prompts slightly affects temporal performance, showing that object-centric prompts are less critical for temporal-only tasks.}
\begin{tabular}{l|cc|cc}
\toprule
\multirow{2}{*}{Models} & \multicolumn{2}{c|}{Charades-STA} & \multicolumn{2}{c}{TVGBench} \\
\cmidrule(lr){2-3} \cmidrule(lr){4-5}
& tIoU@0.3 & tIoU@0.5 & tIoU@0.3 & tIoU@0.5 \\ \midrule
STVG-R1 w. visualprompt & 72.2 & 52.1 & 41.8 & 27.2 \\
STVG-R1 w/o. visialprompt & 73.2 & 52.5 & 42.5 & 27.4 \\
\bottomrule
\end{tabular}
\label{table:timep}
\end{table*}

\subsection{Visualizations of Unseen Category in Object Detector}
\begin{figure}[htbp]
\begin{center}
\includegraphics[width=1\textwidth]{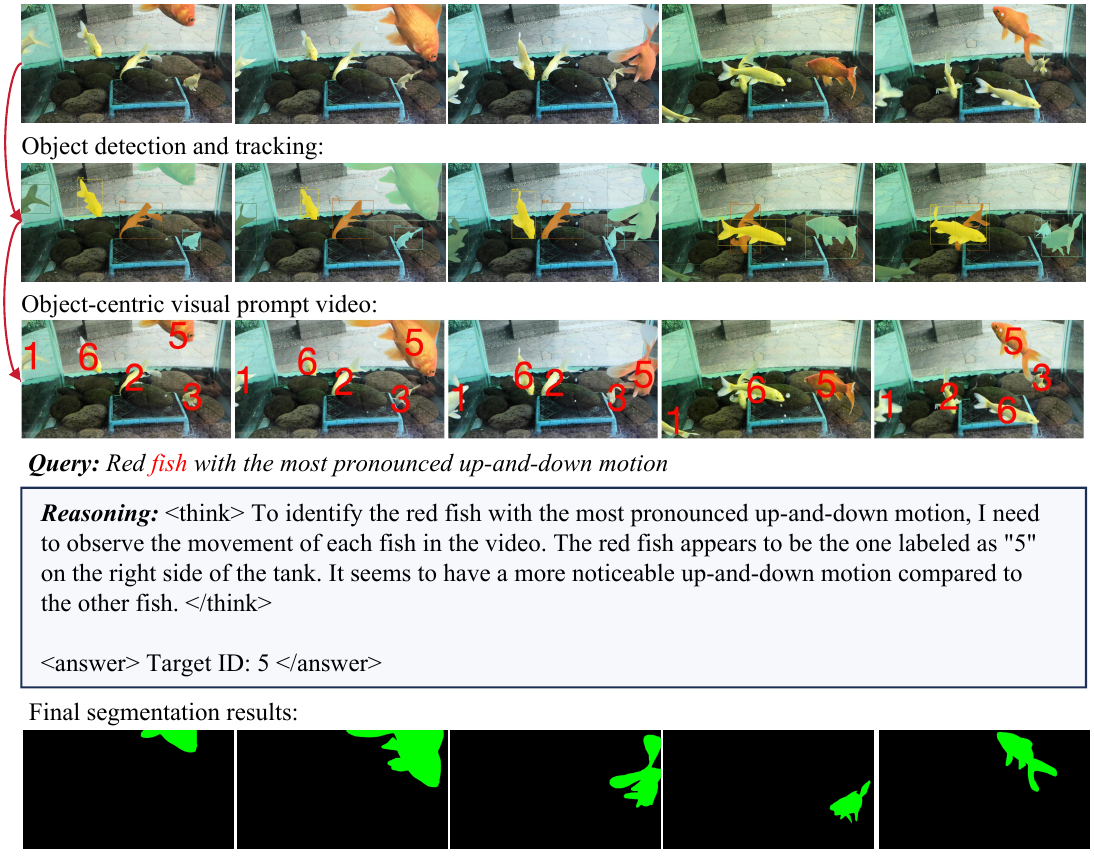}
\end{center}
\setlength{\abovecaptionskip}{0pt}
\caption{Visualizations of unseen category in object detector.}
\label{fig:fish}
\end{figure}

Figure~\ref{fig:fish} presents an example where the queried object (\textit{fish}) is not included in the detector’s taxonomy and is thus misclassified into an incorrect category (\textit{bird}). Nevertheless, our framework assigns a consistent ID and correctly localizes the target instance. This demonstrates that category misclassification does not affect the effectiveness of our approach.

\newpage

\subsection{ID-Repair mechanism during evaluation}
\label{app:IDrepair}
To improve evaluation robustness, we introduce an ID-repair mechanism that corrects missing or inconsistent instance IDs inside the model-predicted temporal segment. Although the detector and SAM2 tracker are generally reliable, occasional ID fragmentation or missed detections can occur, leading to incomplete ID sequences. The ID-repair mechanism algorithm is provided below.

\begin{algorithm}[htbp]
\caption{ID-Repair mechanism during evaluation}
\label{alg:id_repair}
\begin{algorithmic}[1]
\REQUIRE Predicted temporal segment $[t_s, t_e]$; detected boxes $\{B_t\}$ and IDs $\{ID_t\}$ for all frames $t$; target instance ID $ID^*$.
\STATE Initialize ID-correction set $A = \emptyset$
\STATE Let $F = \{t_s, \dots, t_e\}$
\FOR{each frame $t \in F$}
    \IF{$ID^* \in ID_t$}
        \STATE \textbf{continue}
    \ENDIF

    \STATE \textbf{Apply ID correction using $A$}
    \FOR{each $old\_id \in A$}
        \IF{$old\_id \in ID_t$}
            \STATE Replace $old\_id$ with $ID^*$ in $ID_t$
        \ENDIF
    \ENDFOR
    \IF{$ID^* \in ID_t$} 
        \STATE \textbf{continue} 
    \ENDIF

    \STATE Find nearest frame $t_{\text{ref}} \in F$ where $ID^* \in ID_{t_{\text{ref}}}$
    \IF{$t_{\text{ref}}$ exists}
        \STATE Let $b_{\text{ref}}$ be the box of $ID^*$ in $B_{t_{\text{ref}}}$
        \STATE Compute IoU between $b_{\text{ref}}$ and each $b \in B_t$
        \STATE Find $b_{\text{best}}$ with the highest IoU to $b_{\text{ref}}$
        \IF{
            $\operatorname{IoU}(b_{\text{ref}}, b_{\text{best}}) \ge 0.4$
            \OR
            $\displaystyle 
            \frac{\operatorname{Area}(b_{\text{ref}} \cap b_{\text{best}})}
                 {\min\!\left(\operatorname{Area}(b_{\text{ref}}),\, \operatorname{Area}(b_{\text{best}})\right)}
            \ge 0.6$
        }
            \STATE $old\_id \gets$ ID of $b_{\text{best}}$
            \STATE Replace $old\_id$ with $ID^*$ in $ID_t$
            \STATE Add $old\_id$ to $A$ \COMMENT{persist this correction for later frames}
        \ENDIF
    \ENDIF

    \IF{$ID^* \notin ID_t$}
        \STATE Assign $ID^*$ to the box in $B_t$ with largest area
    \ENDIF
\ENDFOR
\end{algorithmic}
\end{algorithm}

\subsection{Ablation of Visual Prompting Pipeline Components}

Table~\ref{tab:vp_ablation} reports ablations on the major components of the preprocessing pipeline. Periodic re-detection is essential, as removing it significantly degrades vIoU by failing to capture objects that appear after the first frame. In contrast, omitting backward tracking leads to a smaller performance drop, indicating that forward tracking alone can recover most trajectories. Thus, backward tracking can be omitted when stricter runtime efficiency is required.

\begin{table}[htbp]
\centering
\caption{Ablation study of preprocessing components on HCSTVG-v1 test set (\%).}
\label{tab:vp_ablation}
\begin{tabular}{lcccc}
\toprule
Method & m\_tIoU & m\_vIoU & vIoU@0.3 & vIoU@0.5 \\
\midrule
w/o re-detection      & 56.1 & 27.8 & 45.4 & 17.4 \\
w/o backward tracking & 56.8 & 28.4 & 66.2 & 37.1 \\
Full pipeline         & 56.9 & 39.1 & 66.7 & 38.6 \\
\bottomrule
\end{tabular}
\end{table}

\subsection{Visualizations for Additional Downstream Tasks}

To further demonstrate the applicability of object-centric visual prompting beyond STVG, we provide qualitative examples for two downstream tasks: video question answering and multi-person video captioning.

\begin{figure}[htbp]
\begin{center}
\includegraphics[width=1\textwidth]{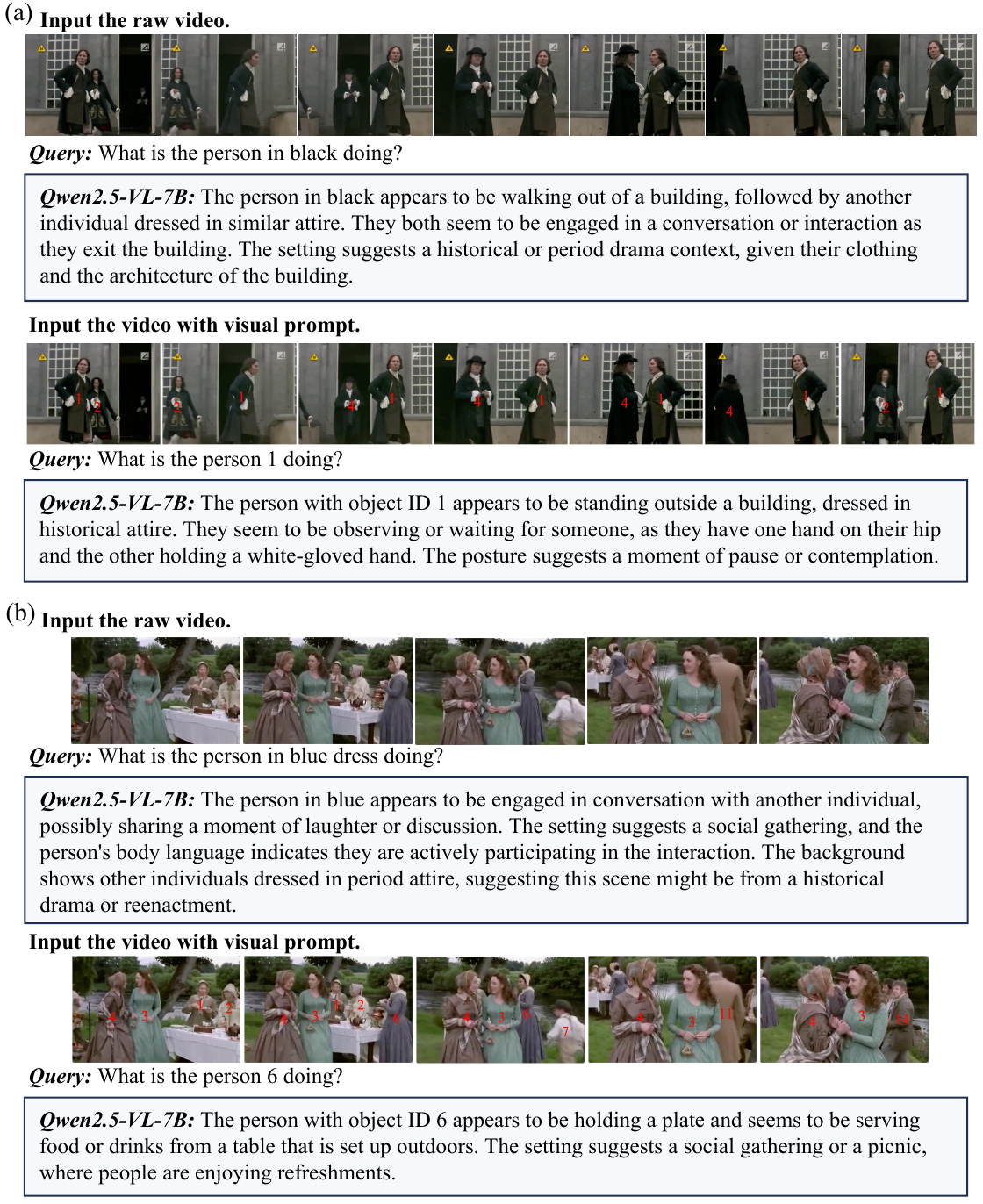}  
\end{center}
\setlength{\abovecaptionskip}{0pt}
\caption{A visualization of video question answering with and without visual prompting.}
\label{fig:vis-vqa}
\end{figure}

\paragraph{Video Question Answering.}
Figure~\ref{fig:vis-vqa} shows cases where the queried entity lacks distinctive appearance cues, making it difficult for a general VLM to localize the correct subject. Without visual prompts, the model fails to answer the question. With object-centric prompts (e.g., “Person 1”), the model correctly grounds the target individual and produces the correct response.

\begin{figure}[htbp]
\begin{center}
\includegraphics[width=1\textwidth]{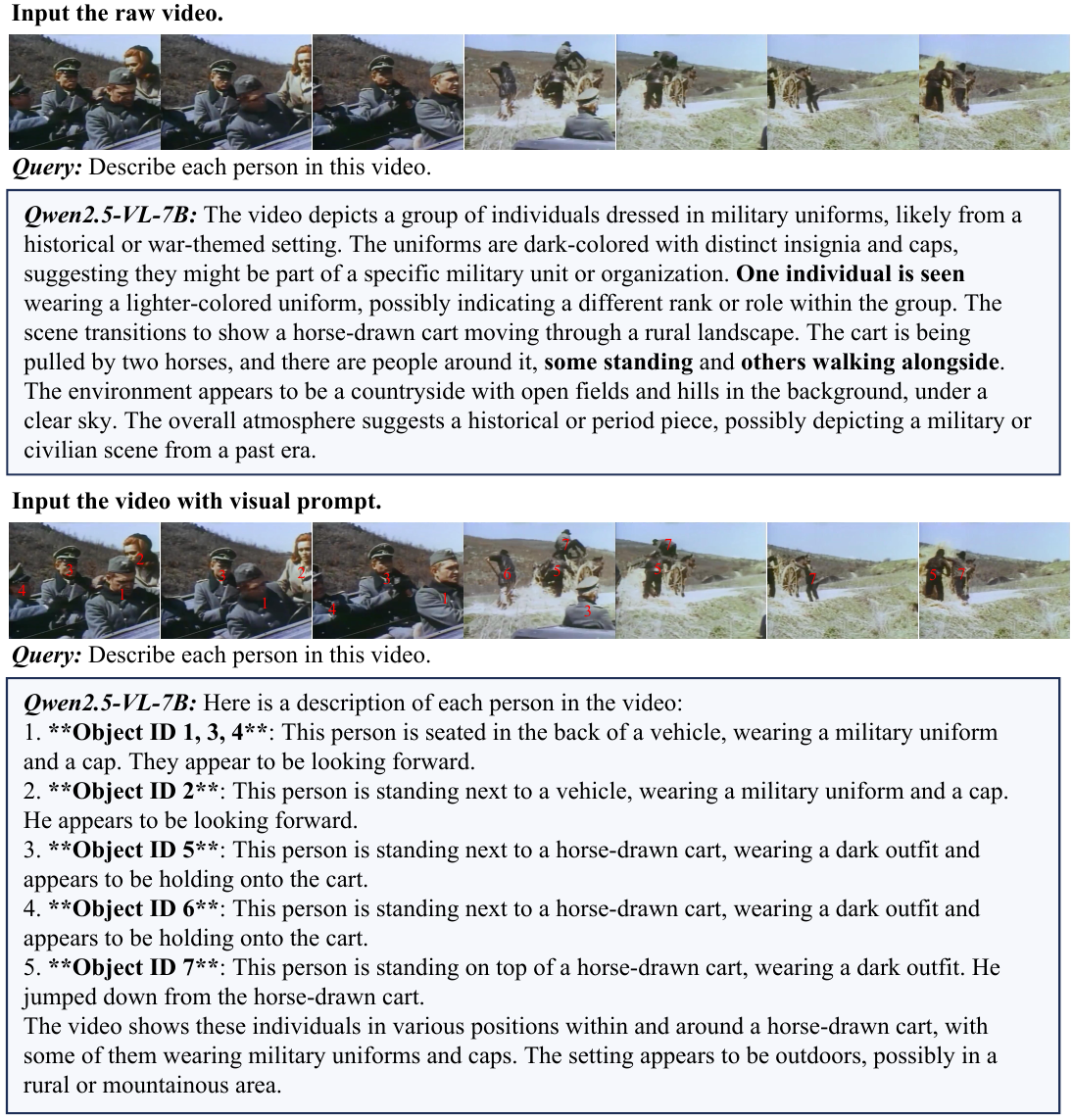}  
\end{center}
\setlength{\abovecaptionskip}{0pt}
\caption{A visualization of multi-person video captioning enhanced by visual prompting.}
\label{fig:vis-caption}
\end{figure}

\newpage
\paragraph{Multi-person Video Captioning.}
Figure~\ref{fig:vis-caption} illustrates how visual prompts benefit video captioning. Without prompts, the model tends to generate a single global caption with ambiguous references. With instance IDs, the model produces entity-specific descriptions for each individual.

\newpage
\subsection{Additional Qualitative Results on Diverse Video Domains}
\begin{figure}[htbp]
\begin{center}
\includegraphics[width=1\textwidth]{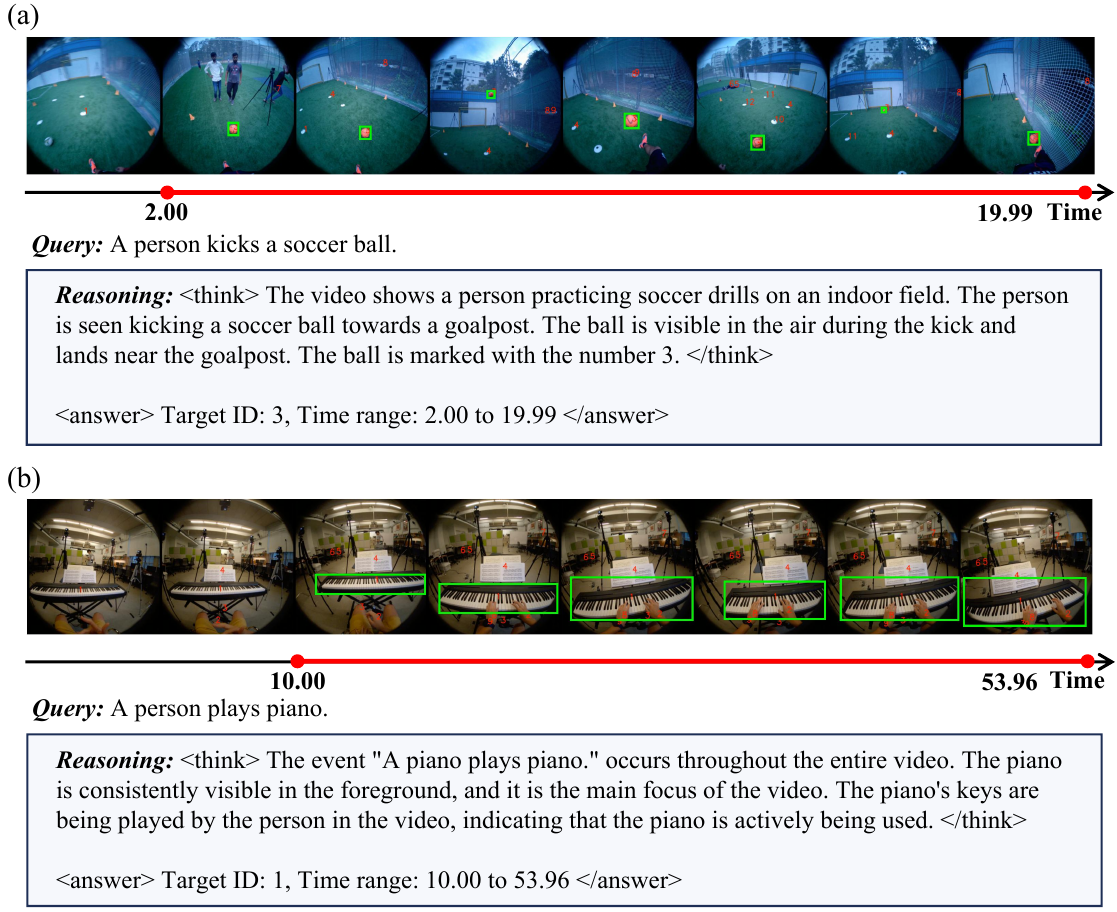}
\end{center}
\setlength{\abovecaptionskip}{0pt}
\caption{Qualitative results on ego-centric videos from Ego4D \cite{grauman2022ego4d}.}
\label{fig:ego4d}
\end{figure}

\begin{figure}[htbp]
\begin{center}
\includegraphics[width=1\textwidth]{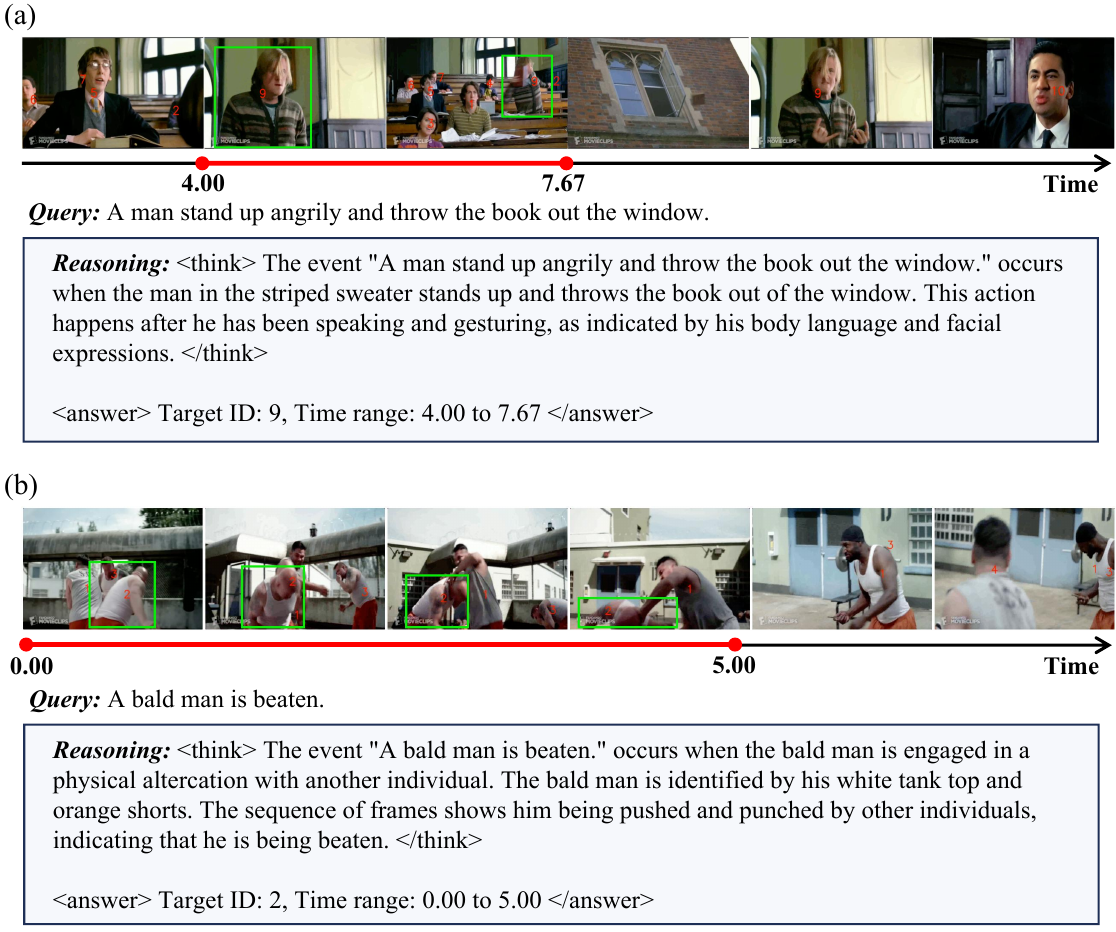}
\end{center}
\setlength{\abovecaptionskip}{0pt}
\caption{Qualitative results on movie-style videos from Grounded-VIDSitu \cite{khan2022grounded}.}
\label{fig:movie}
\end{figure}

To further assess the generalization ability of our approach, we conduct qualitative evaluations on videos drawn from two distinct domains: ego-centric videos from Ego4D \cite{grauman2022ego4d} and movie videos from Grounded-VIDSitu \cite{khan2022grounded}. As shown in Figures~\ref{fig:ego4d} and \ref{fig:movie}, our method maintains stable instance grounding performance despite challenges such as rapid camera motion, complex scene composition, and multiple interacting entities.

\newpage

\subsection{Reward Ablation}
We analyze the effect of the format reward ($r_f$) and a coupled reward variant. All experiments are conducted on HCSTVG-v1.

\begin{figure}[htbp]
    \centering
    \includegraphics[width=0.7\textwidth]{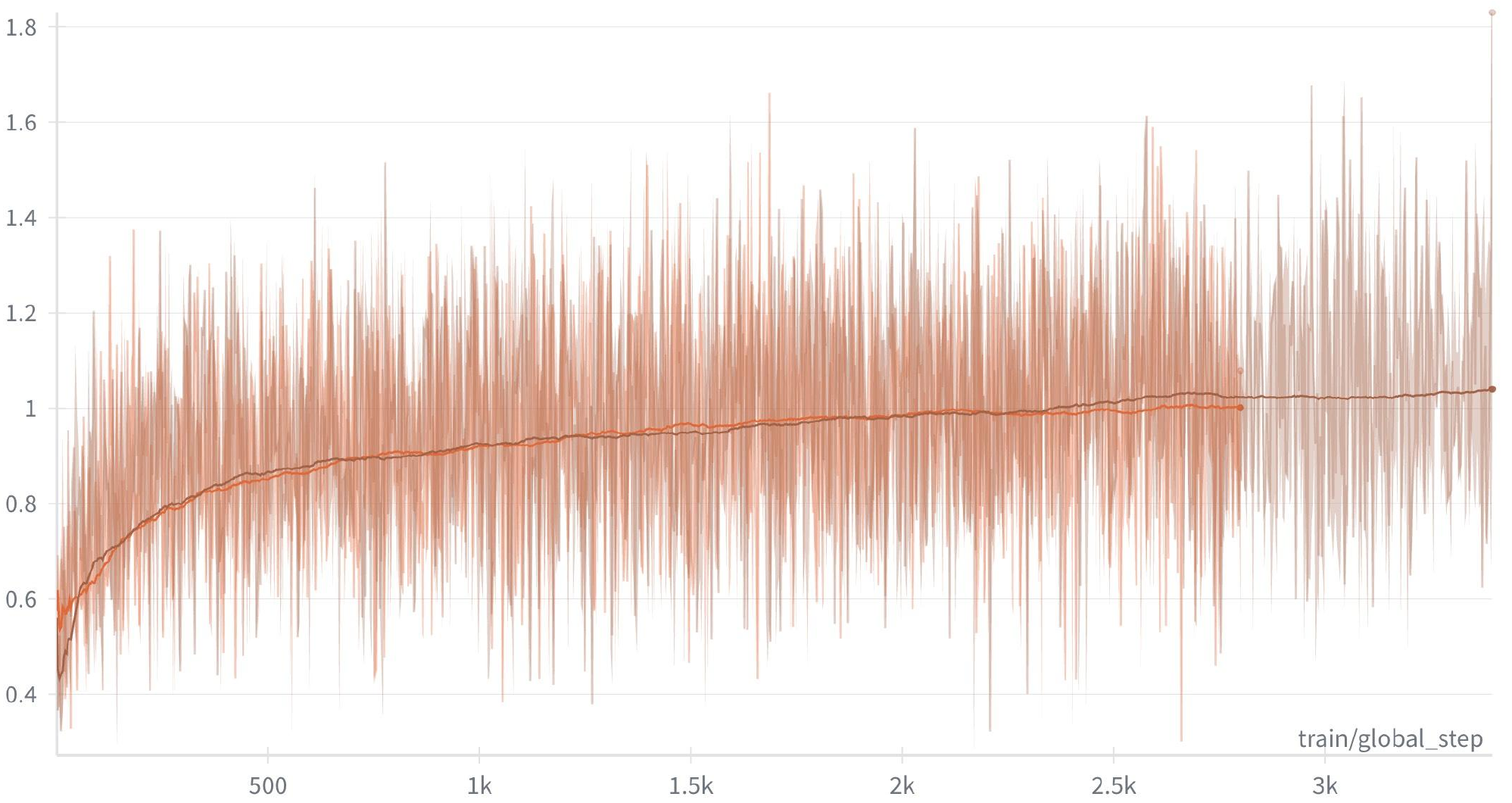} 
    \caption{Training curves with and without the format reward $r_f$. The \textcolor{brown}{brown curve} corresponds to the full reward setting, while the \textcolor{orange}{orange curve} corresponds to the model trained without $r_f$.}
    \label{fig:reward_curve_rf}
\end{figure}

\paragraph{Effect of removing $r_f$.}
The training curves (Fig.~\ref{fig:reward_curve_rf}) indicate that removing $r_f$ produces nearly identical optimization dynamics. Since Qwen2.5-VL naturally supports \texttt{<think>} and \texttt{<answer>} tokens, the marginal benefit of the simple format constraint is limited.

\begin{figure}[htbp]
    \centering
    \includegraphics[width=0.7\textwidth]{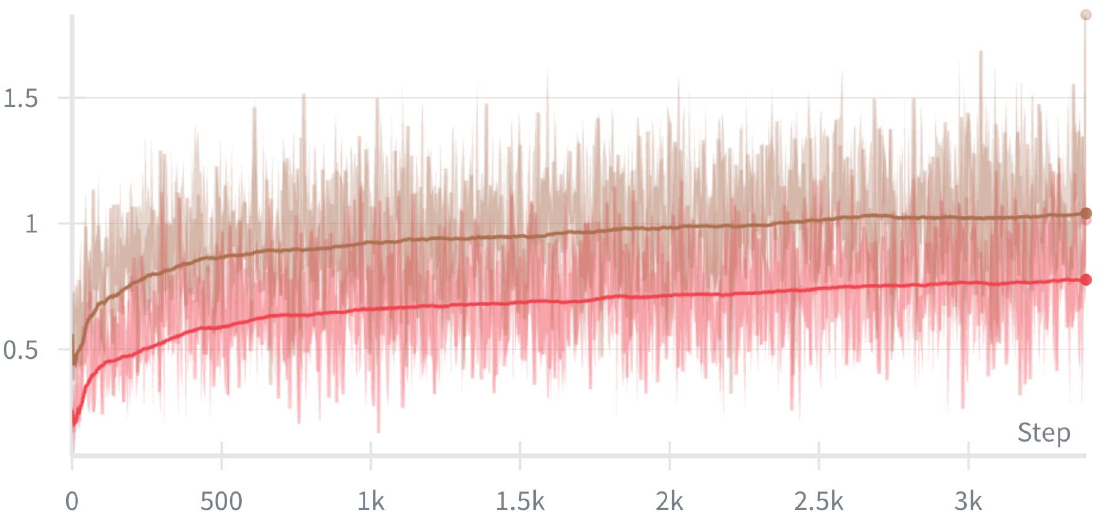}
    \caption{
    Training curves comparing the decoupled \textcolor{brown}{brown reward} and the coupled \textcolor{red}{red reward}.}
    \label{fig:reward_curve_coupled}
\end{figure}

\paragraph{Coupled reward variant.}
The training curves of coupled spatial reward and decoupled spatial reward are as shown in Fig.~\ref{fig:reward_curve_coupled}, which indicate that both formulations exhibit nearly identical convergence behavior.

\subsection{The Use of Large Language Models (LLMs)}
We employed large language models for language polishing to improve the clarity and readability of the manuscript. Specifically, LLMs were used to refine grammar, adjust sentence structure, and enhance overall flow, without altering the technical details and experimental results.

\end{document}